\pdfoutput=1

\documentclass[11pt]{article}

\usepackage{acl}

\usepackage{times}
\usepackage{latexsym}

\usepackage[T1]{fontenc}

\usepackage[utf8]{inputenc}

\usepackage{microtype}
\usepackage{amsmath}

\usepackage{enumitem}
\usepackage{adjustbox}

\newcommand\ky[1]{\textcolor{blue}{#1}}

\usepackage{kotex}
\usepackage{adjustbox}
\usepackage{booktabs}
\usepackage{tikz}
\usepackage{listings}
\usepackage{color}
\usepackage{float}
\restylefloat{table}
\usepackage{xcolor}
\usepackage[linesnumbered,ruled,vlined]{algorithm2e}

\usepackage{verbatim}
\usepackage{multirow}
\usepackage{multicol}
\usepackage{makecell}
\usepackage{tabularx}
\usepackage{amsfonts}
\usepackage{graphicx}
\usepackage[normalem]{ulem}

\definecolor{dkgreen}{rgb}{0,0.6,0}
\definecolor{gray}{rgb}{0.5,0.5,0.5}
\definecolor{mauve}{rgb}{0.58,0,0.82}

\newcommand{\advsub}[2]{
${\text{\sout{#1}}\text{[}\textcolor{red}{\textit{\textbf{#2}}}\text{]}}$
}
\title{Detection of Word Adversarial Examples in Text Classification: \\Benchmark and Baseline via Robust Density Estimation}

\author{
  KiYoon Yoo \\
  Seoul National University \\
  \texttt{961230@snu.ac.kr} \\\And
  Jangho Kim \\
  NAVER WEBTOON AI  \\
  \texttt{jangho.kim@webtoonscorp.com} \\\AND
  Jiho Jang \\
  Seoul National University \\
  \texttt{geographic@snu.ac.kr} \\\And
  Nojun Kwak \\
  Seoul National University \\
  \texttt{nojunk@snu.ac.kr} \\
  }

\begin{document}
\maketitle
\begin{abstract}
Word-level adversarial attacks have shown success in NLP models, drastically decreasing the performance of transformer-based models in recent years. As a countermeasure, adversarial defense has been explored, but relatively few efforts have been made to detect adversarial examples. 
However, detecting adversarial examples may be crucial for automated tasks (e.g. review sentiment analysis) that wish to amass information about a certain population and additionally be a step towards a robust defense system. To this end, we release a dataset for four popular attack methods on four datasets and four models to encourage further research in this field. Along with it, we propose a competitive baseline based on density estimation that has the highest \textsc{auc} on 29 out of 30 dataset-attack-model combinations.\footnote{https://github.com/anoymous92874838/text-adv-detection}
\end{abstract}

\section{Introduction}
Adversarial examples in NLP refer to seemingly innocent texts that alter the model prediction to a desired output, yet remain imperceptible to humans. In recent years, word-level adversarial attacks have shown success in NLP models, drastically decreasing the performance of transformer-based models in sentence classification tasks with increasingly smaller perturbation rate \citep{jin2020bert, li2020bert, garg2020bae, ren2019generating}. In the image domain, two main lines of research exist to counteract adversarial attacks : adversarial example \textit{detection} and \textit{defense}. The goal of detection is to discriminate an adversarial input from a normal input, whereas adversarial defense intends to predict the correct output of the adversarial input. While works defending these attacks have shown some progress in NLP \citep{zhou2021defense, keller-etal-2021-bert, jones2020robust}, only few efforts have been made in techniques for the sole purpose of detection.   

However, detecting adversarial examples may be as crucial as defending them in certain applications, in which alerting the victim of an existence of adversarial samples suffices. For instance, models used for automation of tasks (e.g. review sentiment analysis, news headline classification, etc) are adopted to efficiently gain information about the true data-generating population (e.g. consumers, news media, etc), rather than the adversary. For such applications, attaining outputs of an adversarial input - whether correct or not - may turn out to be harmful to the system. Accordingly, the \textit{discard-rather-than-correct strategy} which simply discards the detected adversarial input would be a good countermeasure. Moreover, being able to detect adversarial examples may be a step towards building a more robust defense model as the popular defense paradigm, adversarial training, usually suffers from degraded performance on normal inputs \citep{bao2021defending}. With a competent detection system, the normal and adversarial inputs can be processed by two separate mechanisms as proposed by \citet{zhou2019learning}.

\begin{figure}
    \centering
    \includegraphics[width=0.47\textwidth]{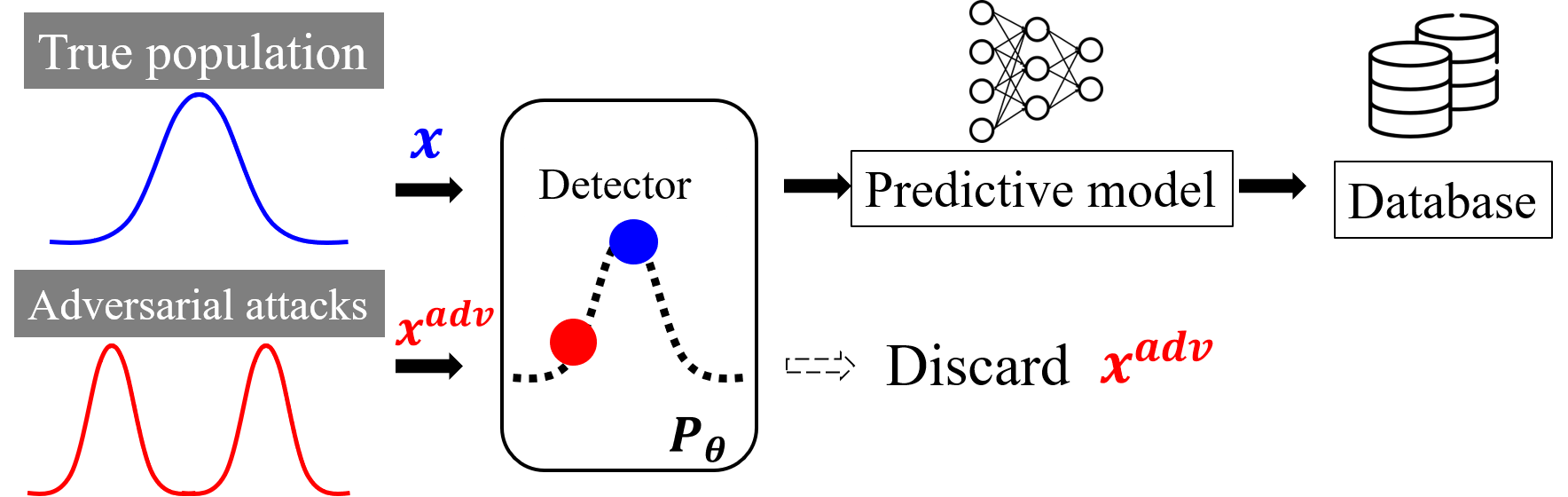}
    \caption{Schematic Diagram of our adversarial detection Framework. We propose a density estimation model to detect adversarial samples.} 
    \label{fig:intro}
    \vspace{-3mm}
\end{figure}

\begin{table*}[t!]
\centering
{\small %
\begin{tabular*}{1.0\textwidth}{llccc}
\hline
    \textbf{Method} & \textbf{Summary} & \textbf{\makecell{Require\\Train Data?}} & \textbf{\makecell{Require\\Val. Data?}} & \textbf{Target Attacks}$^\dagger$ \\
    \hline
    \footnotesize{RDE (Ours)} & \footnotesize{Feature-based density estimation} &  & & Token-level  \\
    \footnotesize{{\citet[FGWS]{mozes2021frequency}}} & \footnotesize{Word frequency-based} &  & \checkmark & Word-level  \\
    \footnotesize{{\citet[ADFAR]{bao2021defending}}} & \footnotesize{Learning-based (sentence-level)} & \checkmark  & \checkmark & Word-level  \\
    \footnotesize{{\citet[DARCY]{le2021sweet}}} & \footnotesize{Learning-based (Honeypot)} & \checkmark & \checkmark & \footnotesize{\citet{wallace2019universal}} \\
    \footnotesize{{\citet[DISP]{zhou2019learning}}} & \footnotesize{Learning-based (token-level)} & \checkmark & \checkmark & Token-level \\
    \footnotesize{{\citet{pruthi2019combating}}} & \footnotesize{Learning-based (Semi-character RNN)} & \checkmark & \checkmark & Char-level \\
    
\hline
\end{tabular*}}
\caption{Key chracteristics of the detection methods in the NLP domain. Requiring training/valdiation data means adversarial samples are needed for training/validation. Token-level encompasses word and character-level. $\dagger$Determined by the experimented types of attacks. Some works can be trivially modified to adjust to a different type of attacks.}
\label{table:method summary}
\vspace{-1mm}
\end{table*}

Many existing works that employ detection as an auxiliary task for defense require adversarial samples for training, which may be a restrictive scenario given the variety of attack methods and sparsity of adversarial samples in the real world. In addition, some works either focus on a single type of attack \citep{le2021sweet} or is limited to character-level attacks \citep{pruthi2019combating}, both of which do not abide the two key constraints (semantics and grammaticality) in order to be imperceptible \citep{morris2020reevaluating}. As opposed to this, carefully crafted word-level adversarial attacks can maintain original semantics and remain unsuspicious to human inspectors.  To encourage further research in this domain, we release a benchmark for word-level adversarial example detection on four attack methods across four NLP models and four text classification datasets. We also propose a simple but effective detection method that utilizes density estimation in the feature space as shown in Fig. \ref{fig:intro} \emph{without any assumption of the attack algorithm or requiring adversarial samples for training or validation}. We summarize the existing works in Table \ref{table:method summary}.

\begin{figure}
\centering
\includegraphics[width=0.35\textwidth]{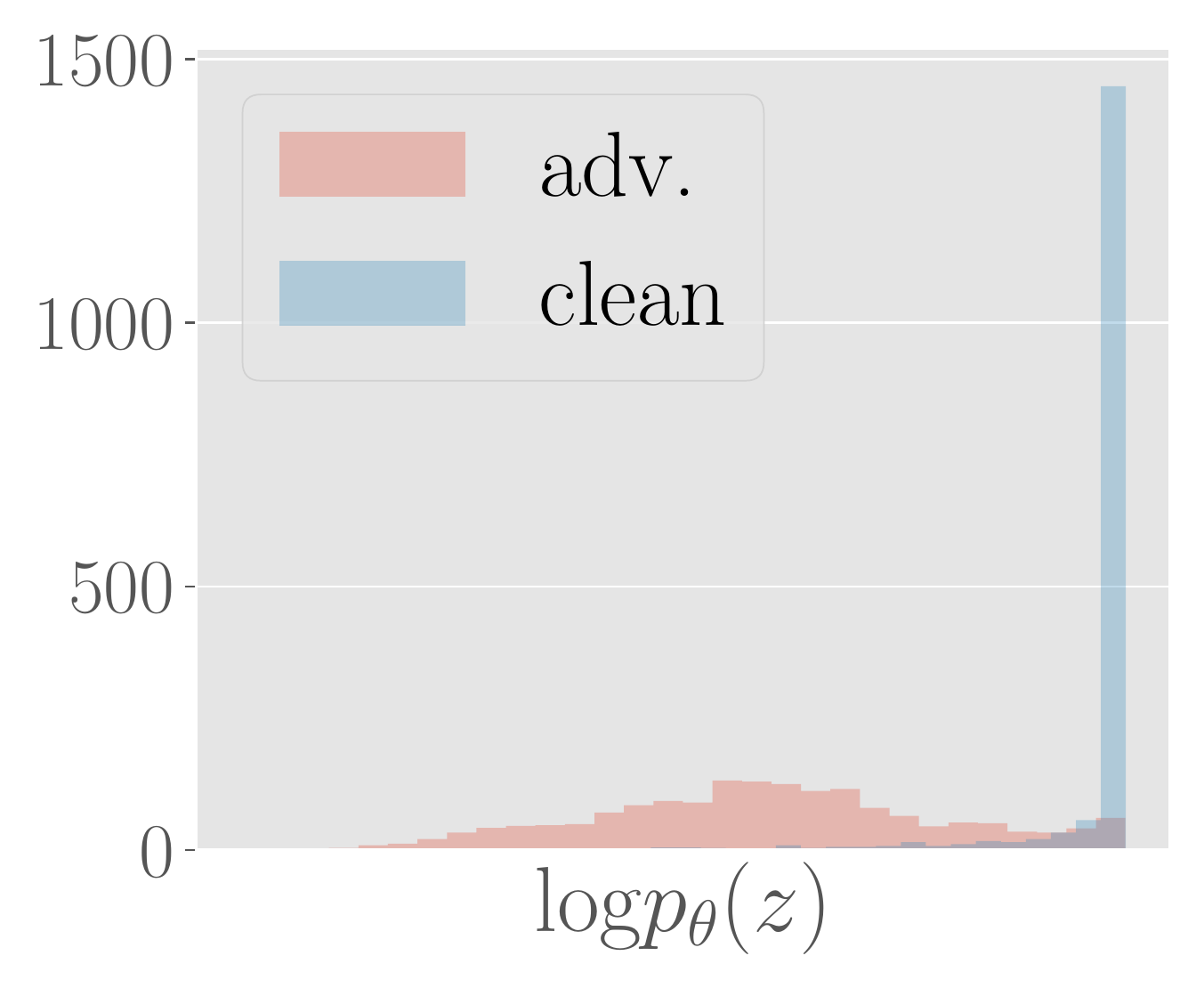}
\caption{Density estimation using our method in (Data-Attack-Model) = (IMDB-(TF-adj)-BERT). Normal samples are peaked at high likelihood region. Adversarial samples tend to have low likelihood.}
\label{fig:confidence}
\end{figure}

As opposed to a recent work \citep{mozes2021frequency}, which relies on word frequency to assess the likelihood of sentence(s), we model the probability density of the entire sentence(s).
To achieve this, we fit a parametric density estimation model to the features obtained from a classification model (e.g. BERT) to yield likelihoods of each sample as shown by Fig. \ref{fig:confidence} inspired by classic works in novelty detection \citep{Bishop1994}, which utilizes generative models to find anomalies.
However, simply fitting a parametric model suffers from curse of dimensionality characterized by (i) sparse data points and spurious features (ii) and rare outliers that hamper accurate estimation. To tackle these issues, we leverage classical techniques in statistical analysis, namely kernel PCA and Minimum Covariance Determinant, for robust density estimation (RDE). 

Our attack-agnostic and model-agnostic detection method achieves the best performance as of \textsc{auc} on 29 out of 30 dataset-attack-model combinations and best performance as of \textsc{tpr, f1 , auc} on 25 of them without any assumption on the attacks.

Our contributions are two-fold:
\vspace{-2mm}

\begin{itemize}[leftmargin=*]
    \item We propose a adversarial detection method that does not require validation sets of each attack method through robust parameter estimation.
    \vspace{-2mm}
    \item We release a dataset for word-level adversarial example detection on 4 attacks, 4 text classification datasets, and 4 models and the source code for experimenting on various experimental protocols.
\end{itemize}
We further provide analysis on a stronger adversary with partial knowledge of the detection method and techniques to counteract the adversary. Last, we investigate the proposed method's applicability on character-level attacks.

\begin{table*}[ht!]
{\small %
\begin{tabular*}{1.0\textwidth}{cccccc}
\hline
    \textbf{Dataset} & \textbf{Topic} & \textbf{Task} & \textbf{Classes} & \makecell{\textbf{Median}\\ \textbf{Length}} & \makecell{\textbf{\# of Test Samples} \\ Original / Generated} \\
    \hline
    IMDB\citep{maas-EtAl:2011:ACL-HLT2011} & movie review & sentiment classification & 2 & 161 & 25K / 10K \\
    AG-News\citep{zhang2015character} &  news headline & topic classification & 4 & 44 & 7.6K / 7.6K \\
    SST-2\citep{socher-etal-2013-recursive} & movie review & sentiment classification & 2 & 16 & 2.7K / 2.7K \\
    YELP\citep{zhang2015character} &restaurant review  & sentiment classification & 2 & 152 & 38K / 5K  \\
\hline
\end{tabular*}}
\caption{Summary of the benchmark dataset. For SST-2, 0.87K held-out validation samples and 1.8K test samples are used.}
\label{table:benchmark}
\vspace{-1mm}
\end{table*}

\section{Preliminaries}
\label{sec2:preliminaries}
\subsection{Adversarial Examples}
Given an input space $\mathcal{X}$, a label space $\mathcal{Y}$, a predictive model $\mathcal{F}:\mathcal{X}\rightarrow \mathcal{Y}$, and an oracle model $\mathcal{F^{*}:\mathcal{X}\rightarrow \mathcal{Y}}$, a successful adversarial example $x_\text{adv}$ of an input $x \in \mathcal{X}$ satisfies the following: 

\begin{equation}
\begin{gathered}
    \label{Eq1: adversarial example}
    \mathcal{F^{*}}(x) = \mathcal{F}(x) \neq \mathcal{F}(x_\text{adv}),\\ 
    C_{i}(x, x_\text{adv})=1\text{ for } i \in \{1, \dots , c\}
\end{gathered}
\end{equation}
where $C_i$ is an indicator function for the  $i$-th constraint between the perturbed text and the original text, which is 1 when the two texts are indistinguishable with respect to the constraint. The constraints vary from attack algorithms and is crucial for maintaining the original semantics while providing an adequate search space. For instance, \citet{jin2020bert} ensure that the embedding of the two sentences have a cosine similarity larger than 0.5 using the Universal Sentence Encoder \citep[USE]{cer2018universal}.

\subsection{Detecting Adversarial Examples}
\label{subsec2.2:detecting}
For the purpose of detecting adversarial examples, a dataset, $\mathcal{D}$, consisting of clean samples ($\mathcal{D}_{\text{clean}}$) and adversarial samples ($\mathcal{D}_{\text{adv}}$) is required.
However, how the dataset should be configured has rarely been discussed in detail and the exact implementation varies by works. Here we discuss two main configurations used in the literature. We denote the test set as $\mathcal{X}_{t}$ and the correctly classified test set as $\mathcal{X}_{c}\subset\mathcal{X}_{t}$.

\vspace{-2mm}
\begin{itemize}[leftmargin = *]
    \item[$\bullet$] Scenario 1 : Sample disjoint subsets $\mathcal{S}_1$, $\mathcal{S}_2\subset\mathcal{X}_{t}$. For the correctly classified examples of $\mathcal{S}_1$, adversarial attacks are generated and the successful examples form $\mathcal{D}_{\text{adv}}$. $\mathcal{D}_{\text{clean}}$ is formed from $\mathcal{S}_2$. 
    \vspace{-2mm}
    \item[$\bullet$] Scenario 2 : Sample subset $\mathcal{S} \subset \mathcal{X}_t$. For the correctly classified examples of $\mathcal{S}$, adversarial attacks are generated and the successful examples form $\mathcal{D}_{\text{adv}}$.  $\mathcal{D}_{\text{clean}}$ is formed from $\mathcal{S}$. 
\end{itemize}

Scenario 1 provides more flexibility in choosing the ratio between adversarial samples and clean samples, while in Scenario 2 this is determined by the attack success rate and task accuracy. For instance, an attack with a low success rate will have a low adversarial-to-clean sample ratio. In addition, Scenario 2 consists of pairs of adversarial samples and their corresponding clean sample in addition to the incorrect clean samples. A more challenging scenario can be proposed by including failed attacked samples, which may be closer to the real world.

A seminal work \citep{xu2017feature} on adversarial example detection in the image domain assumes the first scenario, whereas existing works in NLP \cite{le2021sweet, mozes2021frequency} only experiment on the second scenario.
Our benchmark framework provides the data and tools for experimenting on both. We provide experiment results on both scenarios.

\section{Method}
\label{sec3:method}
\subsection{Benchmark}
\label{subsec:benchmark}
We generate adversarial examples on 4 models, 4 types of attacks, and 4 sentence classification datasets. Since some attacks \citep{garg2020bae} require hundreds of queries and inference of models per query, vast amount of time is required to create all the adversarial examples (e.g. up to 44 hours for 5,000 examples on the IMDB dataset using TF-adjusted attack). This renders on-the-fly generation and detection of adversarial examples extremely inefficient. Therefore, adversarial examples are created beforehand and sampled according to Section \ref{subsec2.2:detecting}. Four sentence classification datasets (IMDB, AG-News, SST-2, Yelp) are chosen to have diverse topics and length. See Table \ref{table:benchmark} for the summary and the number of generated samples. 

We choose two non-transformer-based models (Word-CNN \citet{kimWordCNN}; LSTM \citet{hochreiter1997long}) and two transformer-based models (RoBERTa \citet{liu2019roberta}; BERT \citet{devlin2018bert}) . Recently, numerous adversarial attacks have been proposed. We choose two widely known attacks called Textfooler \citep[TF]{jin2020bert} and Probability Weighted Word Saliency \citep[PWWS]{ren2019generating} and a recent approach using BERT to generate attacks called BAE \citep{garg2020bae}. Lastly, we also include a variant of TF called TF-adjusted \citep[TF-adj]{morris2020reevaluating}, which enforces a stronger similarity constraint to ensure imperceptibility to humans. All attacks are created using the TextAttack library \citep{morris2020textattack}. 
See Appendix \ref{appendix:attack description} for the summary of attack methods and Appendix \ref{appendix:code} for a code snippet of using our benchmark.

\subsection{Estimating Density and Parameters in Feature Space}
\begin{figure}[t]
\centering
\vspace{-5mm}
\includegraphics[width=0.4\textwidth]{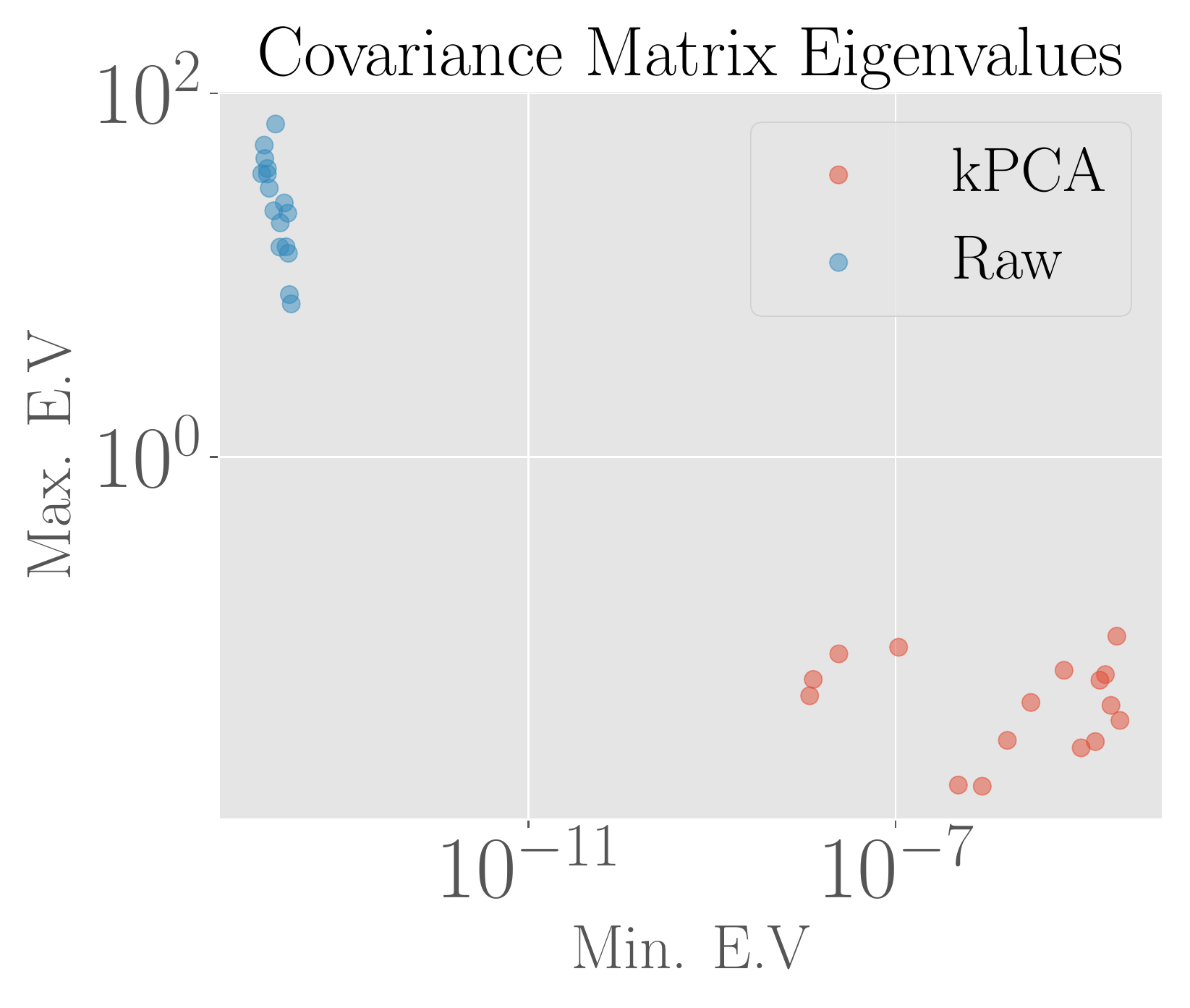}
\caption{Comparisons of the maximum and minimum eigenvalues of the estimated covariance matrices for RoBERTa and BERT across all datasets and classes (20 samples total). Naive estimation (blue) using raw features leads to extremely ill-conditioned matrices while kPCA (red) alleviates this.}
\label{fig:curse1}
\vspace{-3mm}
\end{figure}
Earlier works in novelty detection \citep{Bishop1994} have shown that generative models fitted on normal samples are capable of detecting unseen novel samples (e.g. adversarial samples). Since we can assume that the training samples, which were used to train the victim model of a particular task, are available to the victim party, we can similarly design a generative model that estimates input density. 
However, directly using the inputs is challenging as modeling the probability distribution of raw texts is non-trivial. To bypass this, we fit a parametric density estimation model in the feature space (i.e. penultimate layer of the classification model). Since a neural network learns to extract important features of the inputs to distinguish classes, the features can be regarded as representations of the raw inputs.
For a pre-trained predictive model $\mathcal{F}$, let $z \in \mathcal{Z} \subset \mathbb{R}^{D}$ denote the feature given by the feature extractor $\mathcal{H}: \mathcal{X}\rightarrow\mathcal{Z}$. Then the entire predictive model can be written as the composition of $\mathcal{H}$ and a linear classifier.

Given a generative model $p_{\theta}$ with mean and covariance as parameters $\theta=(\mu,\Sigma)$, we can use the features of the training samples ($\mathcal{X}_{train}$) to estimate the parameters. Then, novel adversarial samples lying in the unobserved feature space are likely to be assigned a low probability, because the generative model only used the normal samples for parameter fitting.
For simplicity, we assume the distributions of the feature $z$ follow a multivariate Gaussian, and thus we model the class conditional probability as $p_{\theta}(z|y=k)\sim N(\mu_k, \Sigma_k) \propto \exp \{ -\frac{1}{2}(z-\mu_k)^T\Sigma_k^{-1} (z-\mu_k)\}$, where $y$ indicates the class of a given task. Then, the maximum likelihood estimate (MLE) is given by the sample mean $\tilde \mu_{\text{MLE}}= \frac{1}{N}\sum_{i=1}^N z_i$ and sample covariance $\tilde \Sigma_{\text{MLE}}= \frac{1}{N-1} \sum_{i=1}^N (z_i-\tilde\mu_{\text{MLE}}) (z_i-\tilde\mu_{\text{MLE}})^T$.

\begin{figure}[t]
\centering
\includegraphics[width=0.43\textwidth]{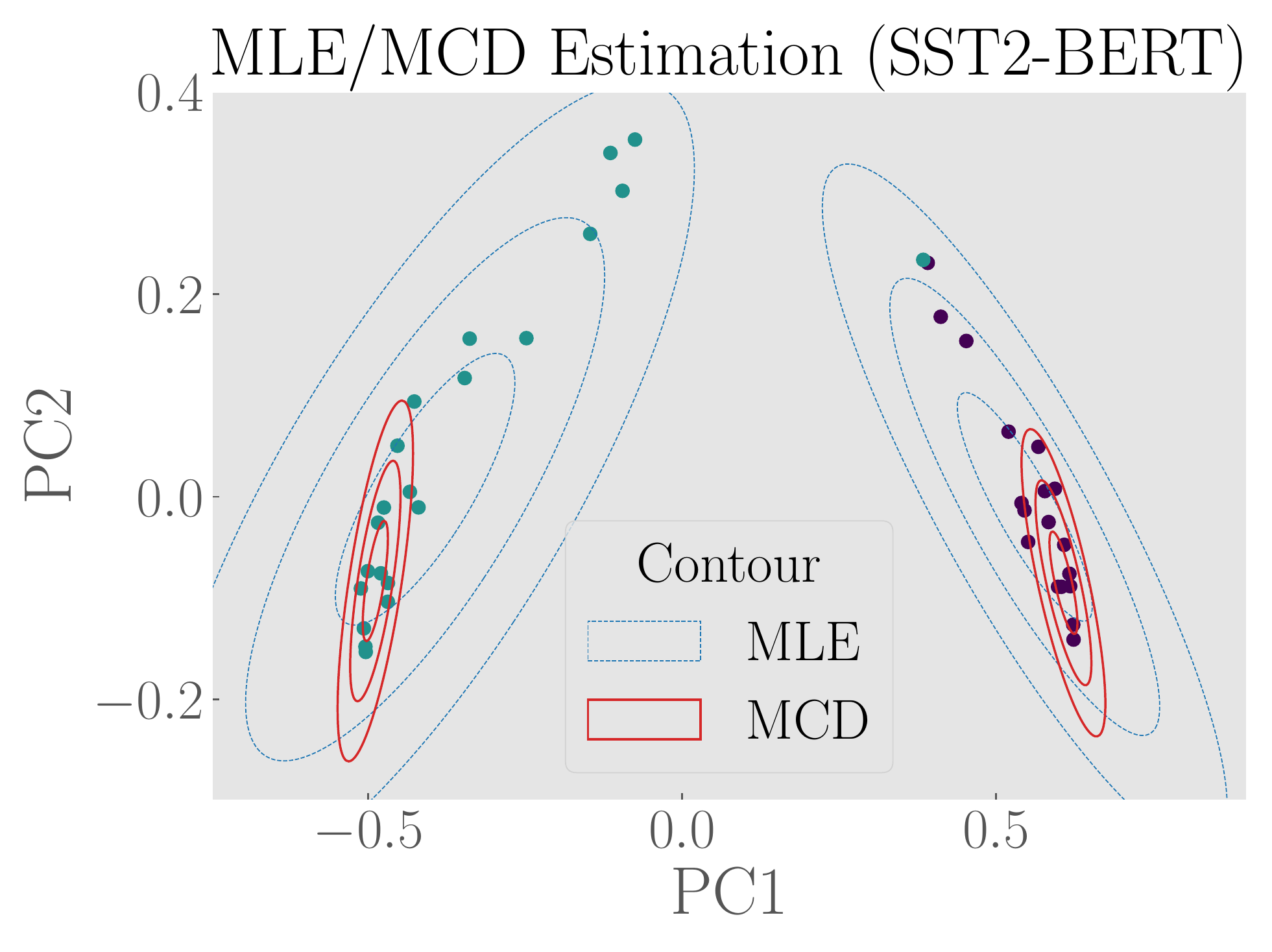}
\caption{Probability-contours of $\tilde\Sigma$ within three standard deviations estimated by MLE and Minimum Covariance Determinant (MCD) of BERT on SST-2 dimensionality reduced by kPCA. Colors of the points indicate class. See Sec. \ref{subsec:3.3} for details.}
\label{fig:curse2}
\vspace{-3mm}
\end{figure}
However, accurate estimation of the parameters is difficult with finite amount of samples especially in high dimensions ($D=768$ for transformer-based models) due to curse of dimensionality, thereby (i) leading to sparse data points and spurious features (ii) and occasional outliers that influence the parameter estimates. 
In Figure \ref{fig:curse1}, we empirically show that the covariance matrices (blue) of BERT and RoBERTa across all models across all datasets are ill-conditioned, demonstrated by the high maximum eigenvalues and extremely small minimum eigenvalues ($\approx 10^{-12}$). Due to this, the precision matrix is abnormally inflated in certain dimensions and prone to numerical errors during inversion. More analysis regarding the upperbound of this error is provided in  Appendix \ref{appendix:parameter estimation}. 

In addition, although we have assumed a Gaussian distribution for convenience, the unknown true distribution may be a more general elliptical distribution with thicker tails. This is observed empirically in Figure \ref{fig:curse2} by visualizing the features into two dimensions by dimensionality reduction. Outliers that are far from the modes of both classes (indicated by color) are present: those that are completely misplaced occasionally exist, while subtle outliers that deviate from the Gaussian distribution assumption are common, which influences the MLE estimation. Thus, to accurately estimate the Gaussian parameters, these outliers should be taken into account. In the following subsection, we tackle these issues through well-known classical techniques from statistical analysis. 

\subsection{RDE using kPCA and Minimum Covariance Determinant}
\label{subsec:3.3}
To address the first issue, we first use kernel PCA \citep[kPCA]{scholkopf1998nonlinear} to select top $P$ orthogonal bases that best explain the variance of the data, thereby reducing redundant features. Given $N$ centered samples $Z_{\text{train}} \in \mathbb{R}^{D \times N} = [z_1, \dots ,z_N]$, a mapping function $\phi : \mathbb{R}^{D} \rightarrow \mathbb{R}^{D'}$,
and its mapping applied to each sample $\Phi(Z_{\text{train}}) \in \mathbb{R}^{D' \times N}$, kPCA projects the data points to the eigenvectors with the $P$ largest eigenvalues of the covariance $\Phi(Z_{\text{train}})\Phi(Z_{\text{train}})^T$\footnote{For simplicity, we assume $\Phi(Z_{\text{train}})$ is centered. When this assumption does not hold, slightly modified approach is taken. See Appendix B of  \citet{scholkopf1998nonlinear} for details.}. Intuitively, this  retains the most meaningful feature dimensions, which explains the data the most, while reducing spurious features and improve stability of inversion by decreasing the condition number as shown in Figure \ref{fig:curse1}. By leveraging non-linear $\phi$, we are able to find meaningful non-linear signals in the features as opposed to standard PCA. We use the radial basis function as our kernel. Comparison of performance with standard PCA is provided in Appendix Table \ref{tab:PCA}. For a complete derivation, please refer to \citet{scholkopf1997kernel}.

However, this does not remove sample-level outliers as shown in Figure \ref{fig:curse2}. Since we have assumed a Gaussian distribution, "peeling off" outliers may be favorable for parameter estimation. A principled way of removing outliers for parameter estimation has been an important research area in multivariate statistics and various methods have been developed for robust covariance estimation \citep{friedman2008sparse, ledoit2004well}. Among them, Minimum Covariance Determinant \citep[MCD]{rousseeuw1984least} finds a subset of $h \leq N$ samples that minimizes the determinant of $\Sigma$.\footnote{Although the possible number of subsets is infeasibly large, \citet{rousseeuw1999fast} propose an iterative method that converges relatively fast for $\approx$ 4000 samples with 100 dimensions.} As the determinant is proportional to the differential entropy of a Gaussian up to a logarithm (shown in Appendix \ref{appendix:MCD}), this results in a robust covariance estimation consisting of centered data points rather than outliers. For a review, see \citet{hubert2018minimum}. Qualitatively, we observe in Figure \ref{fig:curse2} that MLE estimates have their means yanked towards the outliers and that the contours are disoriented (Blue). MCD estimates (Red) focus on the high density clusters, which leads to higher performance as will be shown in the experiments.

In summary, we retain informative features by applying kPCA and obtain robust covariance estiamte by using MCD on the train set. Using the estimated robust parameters, we can evaluate the likelihood of a test sample. We treat those with low likelihood as  novel (adversarial) samples. Our algorithm is shown in Algorithm \ref{alg} in the Appendix. We empirically validate the effectiveness of two techniques and discuss the effect of hyper-parameter $P$ and $h$ in the following sections.

\section{Experiments}
\subsection{Experimental Settings}
We experiment on the four datasets (IMDB, AG-News, SST-2, Yelp) and four attack methods described in Section \ref{subsec:benchmark}. Our experiment is based on BERT and RoBERTa as they are widely used competent models for various tasks. Since SST-2 only has 1.8K test samples, TF-adjusted attack was unable to create an adequate number of successful adversarial samples (e.g. 80 samples out of 1.7K). Omitting experiments for these, there are 30 combinations of dataset-attack-model in total.

In addition, we (i) investigate a potential adversary with partial/full knowledge of the detection method (\S 4.6) and (ii) conduct experiments on a character level attack (Appendix \ref{appendix:char-level}) to demonstrate the applicability of our method. Last, we discuss more realistic scenarios for further study and conduct hyper-parameter and qualitative analysis (\S 4.7).

\subsection{Compared Methods}
We compare our robust density estimation method (\textbf{RDE}) with a recently proposed detection method in NLP called \textbf{FGWS}~\citep{mozes2021frequency} which is a word frequency-based method that assumes that rare words appear more often in adversarial samples. We also verify whether Perplexity (\textbf{PPL}) computed by a language model (GPT-2, \citealt{radford2019language}) is able to distinguish normal and adversarial samples as PPL is often used to compare the fluency of the two samples. \textbf{FGWS} implicitly models the input density via word frequencies, while GPT-2 explicitly computes the conditional probability via an auto-regressive tasks. In addition, we adopt \citet{lee2018simple} denoted as MLE, which is a out-of-distribution detector from the image domain. Similar to RDE, \citet{lee2018simple} fits a Gaussian model using the maximum likelihood estimation (MLE) then trains a logistic regressor using the likelihood scores. Since we assume that adversarial samples are not available for training, we do not train a regression model, but only use the likelihood score. For further details, see Section \ref{sec:related}. We compare MLE with two variants of our method: 
\begin{itemize}[leftmargin=*]
    \item \textbf{RDE(-MCD)} : This is a variant of RDE, in which only kPCA is applied to the features without MCD. The results of applying standard PCA instead of kPCA are reported in Table \ref{tab:PCA} of Appendix.
    \vspace{-3mm}
    \item \textbf{RDE} : After applying kPCA, MCD estimate is used. This is the final proposed robust density estimation incorporating both kPCA and MCD. 
    \vspace{-3mm}
\end{itemize}

\begin{table*}
\vspace{-3mm}
\begin{adjustbox}{width=1.0\textwidth}
\begin{tabular}{c|c||c|c|c||c|c|c||c|c|c||c|c|c}
 \hline
 \multirow{3}{*}{Models} & \multirow{3}{*}{Methods} & \multicolumn{12}{c}{Attacks}\\
 \cline{3-14}
 && \multicolumn{3}{c||}{TF} & \multicolumn{3}{c||}{PWWS}& \multicolumn{3}{c||}{BAE} & \multicolumn{3}{c}{TF-adj}\\
 \cline{3-14}
 && \textsc{tpr} & F1 & \textsc{auc} & \textsc{tpr} & F1 & \textsc{auc} & \textsc{tpr} & F1 & \textsc{auc} & \textsc{tpr} & F1 & \textsc{auc} \\
 \hline 
 && \multicolumn{12}{c}{\Large IMDB} \\ 
 \hline
     \multirow{5}{*}{{BERT}} 
    & PPL &48.7$\pm$0.2 & 61.4$\pm$0.2& 76.9$\pm$0.2&37.8$\pm$0.5 & 51.2$\pm$0.5& 71.7$\pm$0.2&27.0$\pm$0.5 & 39.4$\pm$0.5& 67.3$\pm$0.1&24.5$\pm$0.8 & 36.5$\pm$1.0& 67.9$\pm$0.3 \\ 
    & FGWS &84.6$\pm$0.3 & 87.1$\pm$0.2& 87.1$\pm$0.3&\textbf{88.2}$\pm$0.1 & \textbf{89.0}$\pm$0.0& 90.8$\pm$0.0&62.1$\pm$0.3 & 72.3$\pm$0.2& 70.9$\pm$0.3&72.6$\pm$0.9 & 80.6$\pm$0.9& 78.4$\pm$0.6 \\
    & MLE &86.3$\pm$1.1 & 87.9$\pm$0.7& 94.5$\pm$0.2&75.7$\pm$1.4 & 81.5$\pm$0.9& 92.4$\pm$0.2&81.8$\pm$1.3 & 85.3$\pm$0.8& 93.7$\pm$0.2&88.3$\pm$1.0 & 89.1$\pm$0.6& 95.3$\pm$0.2 \\
    \cline{2-14}  
    & RDE(-MCD) &96.3$\pm$0.3 & 93.4$\pm$0.2& 96.8$\pm$0.1&86.9$\pm$0.9 & 88.3$\pm$0.5& 94.6$\pm$0.2&92.5$\pm$0.5 & 91.4$\pm$0.3& 95.8$\pm$0.2&98.2$\pm$0.2 & 94.6$\pm$0.2& 97.6$\pm$0.2 \\ 
    & RDE & \textbf{96.6}$\pm$0.2 & \textbf{93.5}$\pm$0.1& \textbf{97.7}$\pm$0.2&87.8$\pm$0.4 & 88.8$\pm$0.2& \textbf{95.2}$\pm$0.2&\textbf{93.8}$\pm$0.1 &\textbf{ 92.1}$\pm$0.0& \textbf{96.9}$\pm$0.2& \textbf{98.8}$\pm$0.0 & \textbf{95.0}$\pm$0.1& \textbf{98.7}$\pm$0.2 \\ 
    \cmidrule{1-14}\morecmidrules\cmidrule{1-14}
    \multirow{5}{*}{{\small RoBERTa}} 
    & PPL &47.8$\pm$0.1 & 60.6$\pm$0.1& 78.4$\pm$0.1&43.5$\pm$0.7 & 56.7$\pm$0.6& 76.1$\pm$0.2&25.9$\pm$0.4 & 38.2$\pm$0.5& 67.0$\pm$0.2&26.6$\pm$0.9 & 39.0$\pm$1.1& 69.1$\pm$0.4 \\ 
    & FGWS &85.1$\pm$0.1 & 87.4$\pm$0.1& 88.0$\pm$0.1&92.1$\pm$0.2 & 91.4$\pm$0.2& 93.6$\pm$0.2&61.5$\pm$0.2 & 71.8$\pm$0.1& 70.3$\pm$0.1&69.2$\pm$0.4 & 78.0$\pm$0.1& 75.4$\pm$0.2 \\ 
    & MLE &80.5$\pm$1.0 & 84.5$\pm$0.6& 94.0$\pm$0.2&76.8$\pm$1.3 & 82.2$\pm$0.8& 93.3$\pm$0.2&75.5$\pm$1.5 & 81.4$\pm$0.9& 93.1$\pm$0.3&86.4$\pm$2.3 & 88.0$\pm$1.3& 95.3$\pm$0.7 \\ 
    \cline{2-14} 
    & RDE(-MCD) &98.5$\pm$0.1 & 94.5$\pm$0.1& 97.9$\pm$0.1&95.0$\pm$0.3 & 92.7$\pm$0.2& 96.7$\pm$0.1&\textbf{95.4}$\pm$0.4 &\textbf{93.0}$\pm$0.2& 97.0$\pm$0.2&98.6$\pm$0.4 & 94.8$\pm$0.2& 98.1$\pm$0.4 \\ 
    & RDE &\textbf{98.9}$\pm$0.1 & \textbf{94.7}$\pm$0.0& \textbf{98.6}$\pm$0.1&\textbf{95.2}$\pm$0.1 &\textbf{ 92.8}$\pm$0.1& \textbf{97.2}$\pm$0.1&95.3$\pm$0.2 & 92.9$\pm$0.1& \textbf{97.6}$\pm$0.1& \textbf{98.8}$\pm$0.4 & \textbf{95.9}$\pm$0.6& \textbf{99.0}$\pm$0.2 \\ 

 \hline
 && \multicolumn{12}{c}{\Large AG-News} \\ 
 \hline
 \multirow{5}{*}{{BERT}}  
    & PPL &75.7$\pm$0.4 & 81.6$\pm$0.2& 91.0$\pm$0.2&70.8$\pm$0.7 & 78.3$\pm$0.5& 89.5$\pm$0.2&31.2$\pm$1.3 & 44.2$\pm$1.4& 73.0$\pm$0.8&32.8$\pm$1.8 & 45.9$\pm$1.9& 73.3$\pm$0.8 \\ 
    & FGWS &82.4$\pm$0.6 & 85.7$\pm$0.3& 84.2$\pm$0.7&\textbf{91.0}$\pm$0.1 & \textbf{90.6}$\pm$0.1& 90.8$\pm$0.3&64.3$\pm$0.9 & 73.8$\pm$0.6& 71.3$\pm$0.4&63.8$\pm$1.0 & 74.3$\pm$1.0& 71.9$\pm$0.7 \\ 
    & MLE &77.8$\pm$0.5 & 82.9$\pm$0.3& 93.5$\pm$0.1&70.4$\pm$0.9 & 78.0$\pm$0.6& 92.0$\pm$0.1&72.7$\pm$1.8 & 79.6$\pm$1.2& 92.8$\pm$0.4&71.0$\pm$1.6 & 78.9$\pm$0.9& 92.0$\pm$0.2 \\ 
    \cline{2-14} 
    & RDE(-MCD) &\textbf{96.2}$\pm$0.1 & \textbf{93.3}$\pm$0.0& \textbf{97.1}$\pm$0.1&89.8$\pm$0.8 & 90.0$\pm$0.4& \textbf{95.6}$\pm$0.1&93.2$\pm$0.9 & 92.1$\pm$0.6& 96.2$\pm$0.3&96.6$\pm$1.0 & 93.6$\pm$0.5& 96.0$\pm$0.1 \\ 
    & RDE &95.8$\pm$0.2 & 93.2$\pm$0.1& 96.9$\pm$0.1&88.7$\pm$1.0 & 89.3$\pm$0.6& 95.4$\pm$0.1&\textbf{96.6}$\pm$0.1 & \textbf{93.7}$\pm$0.1& \textbf{96.9}$\pm$0.1&\textbf{98.2}$\pm$0.6 &\textbf{ 95.4}$\pm$0.3& \textbf{97.5}$\pm$0.3 \\ 

    \cmidrule{1-14}\morecmidrules\cmidrule{1-14}
    \multirow{5}{*}{{\small RoBERTa}} 
    & PPL &77.1$\pm$0.5 & 82.4$\pm$0.3& 91.8$\pm$0.1&72.2$\pm$0.8 & 79.3$\pm$0.5& 89.6$\pm$0.2&37.1$\pm$1.4 & 50.4$\pm$1.5& 74.7$\pm$0.3&31.8$\pm$1.3 & 45.3$\pm$1.3& 74.3$\pm$1.3 \\ 
    & FGWS &78.8$\pm$0.5 & 83.5$\pm$0.3& 82.2$\pm$0.2&\textbf{86.6}$\pm$0.4 & \textbf{88.1}$\pm$0.2& 87.9$\pm$0.3&53.3$\pm$3.4 & 65.1$\pm$2.7& 63.5$\pm$2.0&58.9$\pm$3.4 & 69.7$\pm$2.6& 70.1$\pm$0.9 \\ 
    & MLE &82.5$\pm$0.3 & 85.7$\pm$0.2& 94.1$\pm$0.1&78.6$\pm$0.5 & 83.4$\pm$0.3& 92.9$\pm$0.2&68.1$\pm$3.1 & 76.3$\pm$2.2& 91.5$\pm$0.7&65.0$\pm$2.3 & 74.4$\pm$1.7& 91.2$\pm$0.2 \\ 
    \cline{2-14} 
    & RDE(-MCD) &90.5$\pm$0.5 & 90.3$\pm$0.3& \textbf{96.1}$\pm$0.1&84.1$\pm$1.2 & 86.6$\pm$0.7& \textbf{94.8}$\pm$0.2&77.8$\pm$4.1 & 82.6$\pm$2.6& 93.9$\pm$0.5&82.6$\pm$2.7 & 85.9$\pm$1.5& 94.5$\pm$0.4 \\ 
    & RDE &\textbf{92.9}$\pm$0.3 & \textbf{91.6}$\pm$0.2& 95.7$\pm$0.1&84.5$\pm$0.8 & 86.9$\pm$0.5& 93.9$\pm$0.2&\textbf{89.3}$\pm$2.3 & \textbf{89.6}$\pm$1.3& \textbf{95.3}$\pm$0.5&\textbf{94.4}$\pm$0.7 & \textbf{92.6}$\pm$0.4& \textbf{96.0}$\pm$0.3 \\ 

 \hline
 && \multicolumn{12}{c}{\Large SST-2} \\ 
 \hline
 \multirow{5}{*}{{BERT}}  
    & PPL &31.7$\pm$0.6 & 44.8$\pm$0.6& 73.1$\pm$0.3&29.2$\pm$1.3 & 41.9$\pm$1.4& 73.4$\pm$0.4&22.2$\pm$1.6 & 33.5$\pm$2.0& 67.0$\pm$0.5 \\ 
    & FGWS &60.8$\pm$0.4 & 72.3$\pm$0.3& 73.6$\pm$0.3&\textbf{79.9}$\pm$0.6 & \textbf{84.9}$\pm$0.4& \textbf{86.7}$\pm$0.4&34.7$\pm$0.3 & 48.0$\pm$0.3& 60.3$\pm$0.3 \\ 
    & MLE &33.3$\pm$1.3 & 46.5$\pm$1.4& 79.8$\pm$0.5&23.2$\pm$0.4 & 34.8$\pm$0.6& 78.4$\pm$0.3&32.6$\pm$1.3 & 45.8$\pm$1.5& 76.8$\pm$0.6 \\
    \cline{2-11}
    & RDE(-MCD) &61.3$\pm$0.8 & 71.6$\pm$0.6& 86.3$\pm$0.4&46.6$\pm$0.7 & 59.5$\pm$0.6& 84.6$\pm$0.2&45.4$\pm$1.3 & 58.5$\pm$1.1& 80.6$\pm$0.6 \\ 
    & RDE &\textbf{66.1}$\pm$0.8 & \textbf{75.1}$\pm$0.5& \textbf{87.7}$\pm$0.3&54.3$\pm$1.1 & 66.1$\pm$0.9& 86.6$\pm$0.2&\textbf{48.0}$\pm$1.4 & \textbf{60.7}$\pm$1.2& \textbf{81.0}$\pm$0.5 \\ 
    \cmidrule{1-11}\morecmidrules\cmidrule{1-11}
    \multirow{5}{*}{{\small RoBERTa}} 
    & PPL &34.7$\pm$0.7 & 48.0$\pm$0.7& 75.0$\pm$0.5&32.5$\pm$1.6 & 45.5$\pm$1.7& 73.9$\pm$0.5&20.0$\pm$1.3 & 30.8$\pm$1.6& 65.3$\pm$0.4 \\ 
    & FGWS &61.6$\pm$0.2 & 73.0$\pm$0.1& 73.7$\pm$0.1&\textbf{80.8}$\pm$0.2 & \textbf{85.6}$\pm$0.1& 86.4$\pm$0.2&36.1$\pm$1.0 & 49.4$\pm$1.1& 60.0$\pm$0.6 \\ 
    & MLE &44.2$\pm$0.6 & 57.3$\pm$0.5& 84.4$\pm$0.3&33.1$\pm$0.8 & 46.3$\pm$0.8& 81.9$\pm$0.4&37.1$\pm$0.5 & 50.5$\pm$0.5& 77.9$\pm$0.4 \\ 
    \cline{2-11}
    & RDE(-MCD) &63.2$\pm$0.2 & 73.0$\pm$0.1& 87.8$\pm$0.1&53.1$\pm$0.7 & 65.1$\pm$0.6& 85.4$\pm$0.2&45.7$\pm$0.7 & 58.7$\pm$0.7& 79.3$\pm$0.3 \\ 
    & RDE &\textbf{74.1}$\pm$0.3 & \textbf{80.6}$\pm$0.2& \textbf{90.4}$\pm$0.1&67.7$\pm$1.1 & 76.2$\pm$0.8& \textbf{89.1}$\pm$0.0&\textbf{52.0}$\pm$0.3 &\textbf{ 64.3}$\pm$0.3& \textbf{80.6}$\pm$0.1 \\ 
\hline
\end{tabular}
\end{adjustbox}
\caption{Adversarial detection results for BERT and RoBERTa on three datasets on Scenario 1. For all metrics, highers mean better.}
\label{tab:scenario1}
\vspace{-3mm}
\end{table*}

\subsection{Evaluation Metric and Protocol}
Following \citet{xu2017feature}, we report three widely used metrics in adversarial example detection : (1) \textit{True positive rate (\textsc{tpr})} is the fraction of true adversarial samples out of predicted adversarial samples. (2) \textit{F1-score} ({f1}) measures the harmonic mean of precision and recall. Since all compared methods are threshold-based methods, we report \textsc{tpr} at a fixed false positive rate (\textsc{fpr}). (3) \textit{Area under ROC (\textsc{auc})} measures the area under \textsc{tpr} vs. \textsc{fpr} curve. For all three metrics, higher denotes better performance. 

Note that whereas \textsc{auc} considers performance on various \textsc{fpr}'s, \textsc{tpr} and F1 is dependent on one particular \textsc{fpr}. In all our experiments, we fixed \textsc{fpr}$=0.1$, which means $10\%$ of normal samples are predicted to be adversarial samples. This threshold should be chosen depending on the context (i.e. the degree of safety-criticalness). We believe this standard should be elevated as more works are proposed in the future. For IMDB and AG-News, 30\% of the test set is held out as validation set for \citet{mozes2021frequency}. We subsample out of the test set as described in Section \ref{subsec2.2:detecting}. For quantitative analysis, we report the mean and its standard error of three repetitions of random seeds for test/validation split and subsampled samples.

\subsection{Implementation Details}
We choose the feature $z$ to be the output of the last attention layer (before Dropout and fully connected layer) for BERT and RoBERTa. RDE has two main hyper-parameters, namely the number of retained dimensions $P$ of kPCA and the support fraction $h$ of MCD. We fix $P=100$ for all experiments as we observe the performance is not sensitive to $P$. For $h$, we use the default value proposed in the algorithm, which is $\frac{N+P+1}{2N}$. We study the effect of $h$ in Section \ref{subsec:discussion}. All models are pre-trained models provided by TextAttack and both kPCA and MCD are implemented using scikit-learn \citep{scikit-learn}.  We use the radial basis function as our kernel. The time required to estimate the parameters of our method is approximately around 25 seconds. For more details, see Appendix \ref{appendix:implementation detail}.

For FGWS, we use the official implementation\footnote{https://github.com/maximilianmozes/fgws} and use the held-out validation set of each attack to tune the threshold for word frequency $\delta$ as done in the original work. For PPL, we use the HuggingFace \citep{wolf-etal-2020-transformers} implementation of GPT-2 \citep{radford2019language}.

\subsection{Results on Static Adversary}
Table \ref{tab:scenario1} demonstrates the results on three datasets and four attacks. Results on Yelp are presented in Appendix Tab. \ref{tab:yelp} . The highest means out of the four methods are written in bold. Out of the \textbf{30} combinations of dataset-attack-model, RDE achieves the best performance on \textbf{29} of them on \textsc{auc} and \textbf{25} of them for all three metrics, which shows the competitiveness of our simple method. The motivation of our method is verified by the large margin of improvement from MLE in almost all cases. Using MCD estimation also further improves performance except in the few cases of AG-News. Large language models (PPL) are able to distinguish between adversarial samples and normal samples in expectation as shown by the higher-than-random metric, but the performance is inadequate to be used as a detector. 
FGWS generally has higher performance compared to PPL, but is inferior to MLE in most cases. Note the degradation of performance in FGWS for BAE and TF-adj, which are more subtle attacks with stronger constraints, as FGWS relies on the use of rare words. This trend is not observed in feature density-based methods (MLE and RDE). 

FGWS outperforms ours on \textsc{tpr} and F1 in five combinations out of 30, but our method has higher \textsc{auc} on four of them. Interestingly, all five results are from PWWS attacks, which indicates that our method is relatively susceptible to PWWS. Nonetheless, \textsc{auc} still remains fairly high: On IMDB and AG-News, the \textsc{auc}'s are all over 0.9. On the other hand, all methods have degraded performance on SST-2, which may be due to shorter sentence lengths. Some examples of ROC curves are presented in Appendix \ref{appendix:ROC Curve}. Improving detection rate in SST-2 is left as future work.

\subsection{Results on Stronger Adversaries}
\label{stronger adversary}
\noindent\textbf{Adaptive Adversary}\\
In this section, we assume that the adversary is aware of the detection method, but does not have full access to the model parameters. Being conscious of the fact that RDE relies on the discriminative feature space, the adversary does not terminate once the attacked sample reaches the decision boundary, but goes on to generate samples that are closer to the feature space of the incorrect target until a given threshold. Such attacks that increase the number of trials have been similarly applied in \citet{xie2019feature} to create stronger attacks. As shown in Tab. \ref{table:adaptive}, feature-based methods including RDE show considerable degradation in performance against these attacks, while PPL and FGWS have increased performance. All standard errors are less than 0.3.   

\begin{table}[h!]
\begin{adjustbox}{width=\columnwidth,center}
\begin{tabular}{c||cc|cc}
\hline
    & \multicolumn{2}{c}{TF-Strong}          & \multicolumn{2}{c}{PWWS-Strong}              \\
    &\textbf{BERT} & \textbf{RoBERTa} & \textbf{BERT} & \textbf{RoBERTa}      \\
\small{Reference}  & 97.7  &  98.6    &  95.2         &  97.2 \\
\cline{2-5}     
PPL &79.0(+2.1)                 & 84.4(+6.0)         & 73.7 (+2.0)           & 81.7 (+5.6)          \\
FGWS& 87.9(+0.8)            & 90.5(+2.5)             & 92.2(+1.4)        & \textbf{95.3}(+1.7) \\
MLE &93.0(-1.5)             & 90.4(-3.6)             & 90.7(-1.7)            & 90.1(-3.2)               \\
RDE & \underline{94.7}(-3.0) &\underline{92.9}(-5.0) & \underline{92.4}(-2.8) &92.0(-5.2) \\
RDE+& \textbf{95.6}         & \textbf{94.5} &  \textbf{94.2} & \underline{94.5}\\

\hline
\end{tabular}
\end{adjustbox}
\caption{\textsc{AUC}($\Delta$) for character level attack on IMDB on two adaptive attacks.
Reference refers the original RDE against static attacks and $\Delta$ refers to the absolute decrease in performance compared to its respective reference. RDE+ refers to applying RDE after finetuning the features.}
\label{table:adaptive}
\end{table}

However, to combat these types of attacks, the defender can use a few examples of the previously detected static adversarial samples to adjust the feature space. Specifically, the features of the predictive model $\mathcal{F}_\psi$ (e.g. BERT) are finetuned such that the adversarial samples are located near the decision boundary and far from the classes. To achieve this, the parameters $\psi$ of the predictive model are trained such that the entropy of the softmax probability can be maximized. 
\begin{equation}
    \psi_{\mathrm{T}} = \sum_{t=1}^{\mathrm{T}} [\psi_t + \sum_{x_i \sim \mathcal{D}_{adv}}^{b} \nabla \mathcal{H}(\mathcal{F}(x_i; \psi_{t-1}))]
\end{equation}
With only \textbf{T=5} iterations and \textbf{80} adversarial samples ($b$=16), RDE+ is able to recover some of its performance by adjusting the feature space. Since only a few updates are made to the model, the original task performance is negligibly affected (For RoBERTa, 95.1\% is marginally increased to 95.2\%).\footnote{For more details regarding the experiment, see the appendix.} Note that only the previously detected static adversarial samples are used to finetune the feature space not the stronger attacks from the adaptive adversary. In addition, stronger attacks are usually more perceptible as it perturbs more words and cost more queries. For instance, on RoBERTa-TF, the average number of queries to the model per sample increasees from 625 $\rightarrow$ 786.

\begin{figure}[!h]
\vspace{-1mm}
    \centering
    \includegraphics[width=1.0\columnwidth]{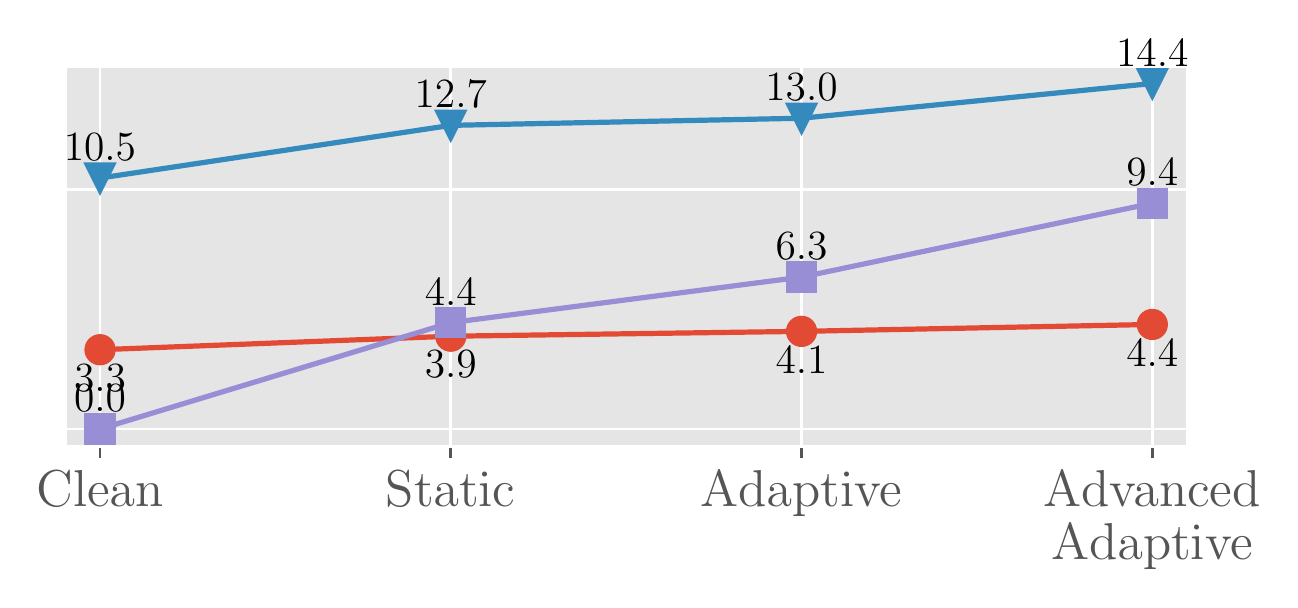}
    \caption{Comparing grammatical error (\textcolor{blue}{Triangle}), PPL (\textcolor{red}{Circle}), and dissimilarity with the original sentence (\textcolor{violet}{Sqaure}) using USE\citep{cer2018universal} } 
    \label{fig:percep-analysis}
    \vspace{-2mm}
\end{figure}

\noindent\textbf{Advanced Adaptive / Oracle Adversary}\\
If the adversary has full access to the model parameters, the adversary can easily generate attacks that can evade the detection method by iteratively attacking the likelihood score of RDE. We generate an even stronger attack than the adaptive adversary (advanced adaptive) such that the attacked sample is completely misclassified to the incorrect target to approximate the oracle adversary. However, we show in Figure \ref{fig:percep-analysis} that such contrived attacks evade RDE at a cost of high perceptibility (grammatical error, PPL, semantic defect), rendering them detectable by human inspectors or other detection methods not reliant on the neural features. 

\subsection{Discussion}
\label{subsec:discussion}
\noindent\textbf{More Realistic Scenarios}
In previous experiments, all failed adversarial attacks were discarded. However, in reality an adversary will probably have no access to the victim model so some attacks will indeed fail to fool the model and have unbalanced clean to adversarial samples. In Appendix \ref{appendix:scenario2}, we discuss these scenarios and provide preliminary results.

\begin{figure}[!h]
\centering
\includegraphics[width=0.48\textwidth]{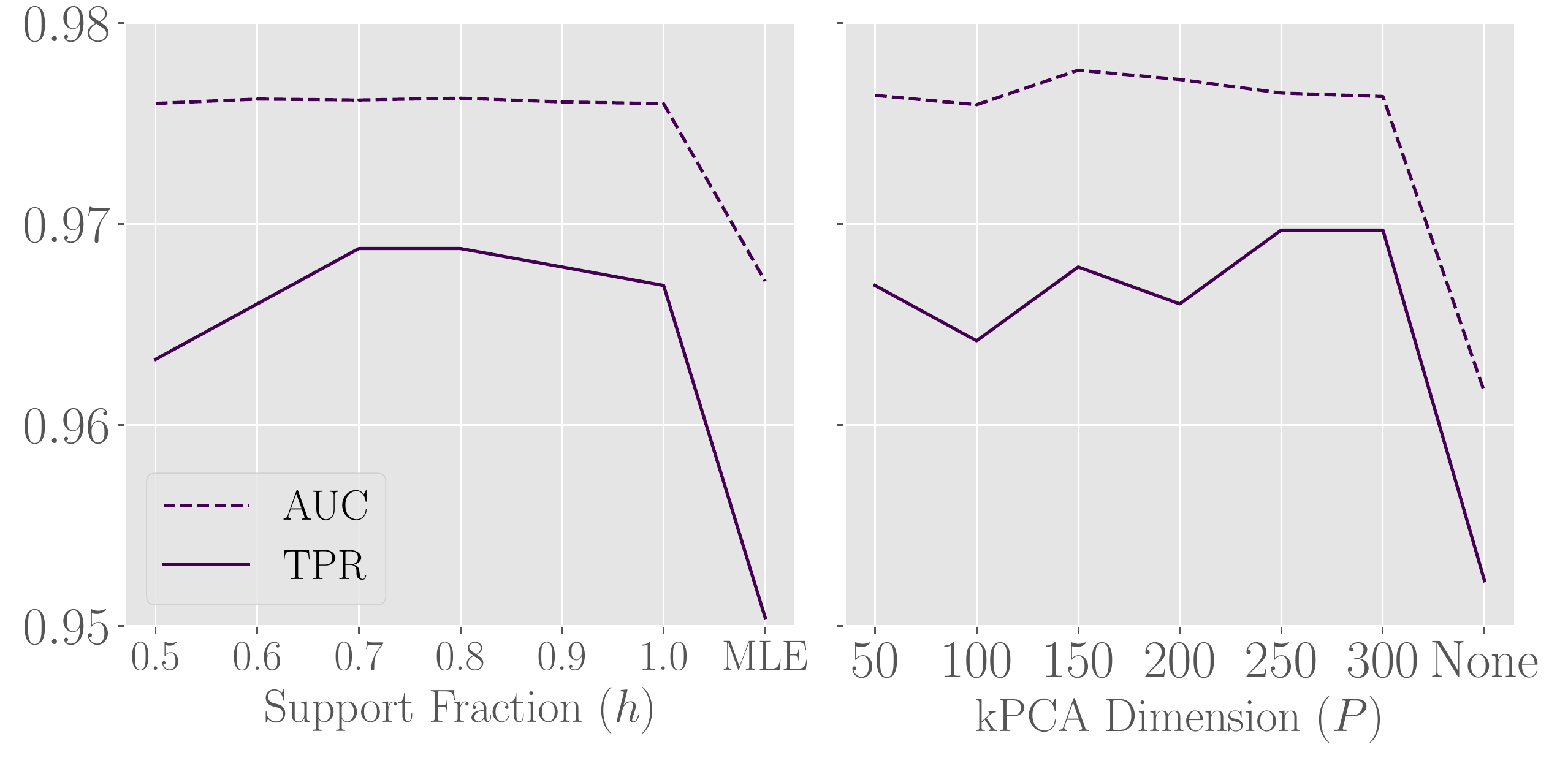}
\caption{Hyperparameter analysis on IMDB-TF-BERT. Wide range of values all outperform the ablated forms and are relatively stable.}
\vspace{3mm}
\label{fig:hyperparam}
\end{figure}

\noindent\textbf{Hyper-parameter Analysis}
Although the two main hyper-parameters, support fraction ($h$) of MCD and the dimension ($P$), were fixed in our experiments, they can be fine-tuned in a separate validation set for optimal performance. We show in Figure \ref{fig:hyperparam} the performance of our method on various ranges of $h$ and $P$ on the validation set of IMDB-TF-BERT combination. 
We set $P=100$ and $h$ to the default value of the algorithm when tuning for the other parameter. 

\begin{figure}[!h]
\vspace{-1mm}
\centering
\includegraphics[width=0.38\textwidth]{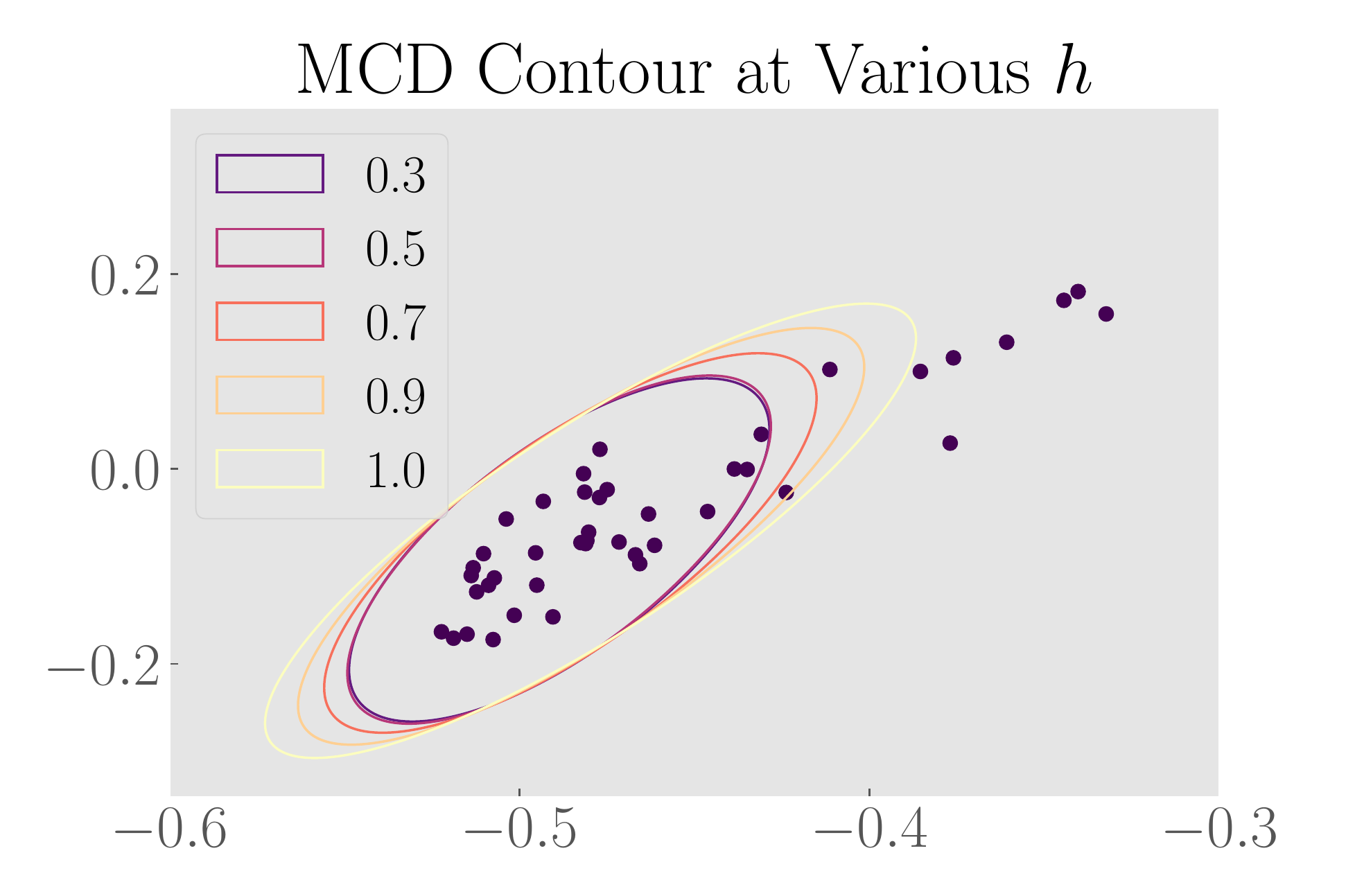}
\caption{Qualitative example of varying support fraction $h$ on SST2-BERT. Each ellipse represent probability contours of three standard deviations. Higher $h$ retains more of the deviating samples and leads to wider contour.}
\label{fig:h_analysis}
\end{figure}

\noindent\textbf{Qualitative Analysis on Support Fraction}
\noindent The support fraction controls the ratio of original samples to be retained by the MCD estimator, thereby controlling the volume of the contour as shown in Figure \ref{fig:h_analysis}. We empirically demonstrated in our experiment that using all samples for parameter estimation may be detrimental for adversarial sample detection.  

\section{Related Works}
\label{sec:related}
Detection of adversarial examples is a well-explored field in the image domain. Earlier works have tackled in various ways such as input transformation \citep{xu2017feature}, statistical analysis \citep{grosse2017statistical}, or training a separate binary classifier \citep{metzen2017detecting}. However, \citet{carlini2017adversarial} has shown that an adversary with partial knowledge of the detector can easily nullify it. Meanwhile, early works in novelty detection have shown that a generative model can detect anomaly samples \citep{Bishop1994}. Following this line of research, \citet{lee2018simple}
have proposed a method to detect out-of-distribution samples by training a logistic regressor on features of a neural network for maximum likelihood estimation.

In the NLP domain, few efforts have been made in detecting word-level adversarial examples. \citet[DISP]{zhou2019learning} utilize a detector trained on adversarial samples for a joint detect-defense system. FGWS \citep{mozes2021frequency} outperforms DISP in detection by building on the observation that attacked samples are composed of rare words on 2 attack methods. \citet{le2021sweet} tackle a particular attack method called UniTrigger \citep{wallace2019universal}, which pre-pends or appends an identical phrase in all sentences. While the performance is impressive, applying this method to other attacks requires significant adjustment due to the distinct characteristics of UniTrigger. Meanwhile, \citet{pruthi2019combating} tackle character-level adversarial examples and compare with spell correctors.  Our work is the first to extensively demonstrate experimental results for 4 popular and recent attack methods on 4 datasets and propose a competitive baseline. We summarize the methods in Tab. \ref{table:method summary}.

\section{Conclusion}
\vspace{-3mm}
We propose a general method and benchmark for adversarial example detection in NLP. Our method RDE does not require training or validation sets for each attack algorithms, yet achieves competitive performance. In the future, a principled coutermeausre for an adversary with partial or full knowledge can be considered for robustness.

\bibliography{anthology}

\begin{thebibliography}{44}
\expandafter\ifx\csname natexlab\endcsname\relax\def\natexlab#1{#1}\fi

\bibitem[{Anderson(1962)}]{anderson1962introduction}
Theodore~Wilbur Anderson. 1962.
\newblock An introduction to multivariate statistical analysis.
\newblock Technical report, Wiley New York.

\bibitem[{Bao et~al.(2021)Bao, Wang, and Zhao}]{bao2021defending}
Rongzhou Bao, Jiayi Wang, and Hai Zhao. 2021.
\newblock Defending pre-trained language models from adversarial word
  substitutions without performance sacrifice.
\newblock \emph{arXiv preprint arXiv:2105.14553}.

\bibitem[{Bishop(1994)}]{Bishop1994}
C.M. Bishop. 1994.
\newblock \href
  {https://digital-library.theiet.org/content/journals/10.1049/ip-vis_19941330}
  {Novelty detection and neural network validation}.
\newblock \emph{IEEE Proceedings - Vision, Image and Signal Processing},
  141:217--222(5).

\bibitem[{Carlini and Wagner(2017)}]{carlini2017adversarial}
Nicholas Carlini and David Wagner. 2017.
\newblock Adversarial examples are not easily detected: Bypassing ten detection
  methods.
\newblock In \emph{Proceedings of the 10th ACM workshop on artificial
  intelligence and security}, pages 3--14.

\bibitem[{Cer et~al.(2018)Cer, Yang, Kong, Hua, Limtiaco, John, Constant,
  Guajardo-C{\'e}spedes, Yuan, Tar et~al.}]{cer2018universal}
Daniel Cer, Yinfei Yang, Sheng-yi Kong, Nan Hua, Nicole Limtiaco, Rhomni~St
  John, Noah Constant, Mario Guajardo-C{\'e}spedes, Steve Yuan, Chris Tar,
  et~al. 2018.
\newblock Universal sentence encoder.
\newblock \emph{arXiv preprint arXiv:1803.11175}.

\bibitem[{Devlin et~al.(2018)Devlin, Chang, Lee, and
  Toutanova}]{devlin2018bert}
Jacob Devlin, Ming-Wei Chang, Kenton Lee, and Kristina Toutanova. 2018.
\newblock Bert: Pre-training of deep bidirectional transformers for language
  understanding.
\newblock \emph{arXiv preprint arXiv:1810.04805}.

\bibitem[{Friedman et~al.(2008)Friedman, Hastie, and
  Tibshirani}]{friedman2008sparse}
Jerome Friedman, Trevor Hastie, and Robert Tibshirani. 2008.
\newblock Sparse inverse covariance estimation with the graphical lasso.
\newblock \emph{Biostatistics}, 9(3):432--441.

\bibitem[{Garg and Ramakrishnan(2020)}]{garg2020bae}
Siddhant Garg and Goutham Ramakrishnan. 2020.
\newblock Bae: Bert-based adversarial examples for text classification.
\newblock In \emph{Proceedings of the 2020 Conference on Empirical Methods in
  Natural Language Processing (EMNLP)}, pages 6174--6181.

\bibitem[{Grosse et~al.(2017)Grosse, Manoharan, Papernot, Backes, and
  McDaniel}]{grosse2017statistical}
Kathrin Grosse, Praveen Manoharan, Nicolas Papernot, Michael Backes, and
  Patrick McDaniel. 2017.
\newblock On the (statistical) detection of adversarial examples.
\newblock \emph{arXiv preprint arXiv:1702.06280}.

\bibitem[{Hochreiter and Schmidhuber(1997)}]{hochreiter1997long}
Sepp Hochreiter and J{\"u}rgen Schmidhuber. 1997.
\newblock Long short-term memory.
\newblock \emph{Neural computation}, 9(8):1735--1780.

\bibitem[{Hubert et~al.(2018)Hubert, Debruyne, and
  Rousseeuw}]{hubert2018minimum}
Mia Hubert, Michiel Debruyne, and Peter~J Rousseeuw. 2018.
\newblock Minimum covariance determinant and extensions.
\newblock \emph{Wiley Interdisciplinary Reviews: Computational Statistics},
  10(3):e1421.

\bibitem[{Jin et~al.(2020)Jin, Jin, Zhou, and Szolovits}]{jin2020bert}
Di~Jin, Zhijing Jin, Joey~Tianyi Zhou, and Peter Szolovits. 2020.
\newblock Is bert really robust? a strong baseline for natural language attack
  on text classification and entailment.
\newblock In \emph{Proceedings of the AAAI conference on artificial
  intelligence}, volume~34, pages 8018--8025.

\bibitem[{Jones et~al.(2020)Jones, Jia, Raghunathan, and
  Liang}]{jones2020robust}
Erik Jones, Robin Jia, Aditi Raghunathan, and Percy Liang. 2020.
\newblock Robust encodings: A framework for combating adversarial typos.
\newblock In \emph{Proceedings of the 58th Annual Meeting of the Association
  for Computational Linguistics}, pages 2752--2765.

\bibitem[{Keller et~al.(2021)Keller, Mackensen, and
  Eger}]{keller-etal-2021-bert}
Yannik Keller, Jan Mackensen, and Steffen Eger. 2021.
\newblock \href {https://doi.org/10.18653/v1/2021.findings-acl.141}
  {{BERT}-defense: A probabilistic model based on {BERT} to combat cognitively
  inspired orthographic adversarial attacks}.
\newblock In \emph{Findings of the Association for Computational Linguistics:
  ACL-IJCNLP 2021}, pages 1616--1629, Online. Association for Computational
  Linguistics.

\bibitem[{Kim(2014)}]{kimWordCNN}
Yoon Kim. 2014.
\newblock \href {http://arxiv.org/abs/1408.5882} {Convolutional neural networks
  for sentence classification}.
\newblock \emph{CoRR}, abs/1408.5882.

\bibitem[{Le et~al.(2021)Le, Park, and Lee}]{le2021sweet}
Thai Le, Noseong Park, and Dongwon Lee. 2021.
\newblock A sweet rabbit hole by darcy: Using honeypots to detect universal
  trigger’s adversarial attacks.
\newblock In \emph{59th Annual Meeting of the Association for Comp. Linguistics
  (ACL)}.

\bibitem[{Ledoit and Wolf(2004)}]{ledoit2004well}
Olivier Ledoit and Michael Wolf. 2004.
\newblock A well-conditioned estimator for large-dimensional covariance
  matrices.
\newblock \emph{Journal of multivariate analysis}, 88(2):365--411.

\bibitem[{Lee et~al.(2018)Lee, Lee, Lee, and Shin}]{lee2018simple}
Kimin Lee, Kibok Lee, Honglak Lee, and Jinwoo Shin. 2018.
\newblock A simple unified framework for detecting out-of-distribution samples
  and adversarial attacks.
\newblock \emph{Advances in neural information processing systems}, 31.

\bibitem[{Li et~al.(2020)Li, Ma, Guo, Xue, and Qiu}]{li2020bert}
Linyang Li, Ruotian Ma, Qipeng Guo, Xiangyang Xue, and Xipeng Qiu. 2020.
\newblock Bert-attack: Adversarial attack against bert using bert.
\newblock In \emph{Proceedings of the 2020 Conference on Empirical Methods in
  Natural Language Processing (EMNLP)}, pages 6193--6202.

\bibitem[{Liu et~al.(2019)Liu, Ott, Goyal, Du, Joshi, Chen, Levy, Lewis,
  Zettlemoyer, and Stoyanov}]{liu2019roberta}
Yinhan Liu, Myle Ott, Naman Goyal, Jingfei Du, Mandar Joshi, Danqi Chen, Omer
  Levy, Mike Lewis, Luke Zettlemoyer, and Veselin Stoyanov. 2019.
\newblock Roberta: A robustly optimized bert pretraining approach.
\newblock \emph{arXiv preprint arXiv:1907.11692}.

\bibitem[{Maas et~al.(2011)Maas, Daly, Pham, Huang, Ng, and
  Potts}]{maas-EtAl:2011:ACL-HLT2011}
Andrew~L. Maas, Raymond~E. Daly, Peter~T. Pham, Dan Huang, Andrew~Y. Ng, and
  Christopher Potts. 2011.
\newblock \href {http://www.aclweb.org/anthology/P11-1015} {Learning word
  vectors for sentiment analysis}.
\newblock In \emph{Proceedings of the 49th Annual Meeting of the Association
  for Computational Linguistics: Human Language Technologies}, pages 142--150,
  Portland, Oregon, USA. Association for Computational Linguistics.

\bibitem[{Metzen et~al.(2017)Metzen, Genewein, Fischer, and
  Bischoff}]{metzen2017detecting}
Jan~Hendrik Metzen, Tim Genewein, Volker Fischer, and Bastian Bischoff. 2017.
\newblock On detecting adversarial perturbations.
\newblock \emph{International Conference on Learning Representations}.

\bibitem[{Morris et~al.(2020{\natexlab{a}})Morris, Lifland, Lanchantin, Ji, and
  Qi}]{morris2020reevaluating}
John Morris, Eli Lifland, Jack Lanchantin, Yangfeng Ji, and Yanjun Qi.
  2020{\natexlab{a}}.
\newblock Reevaluating adversarial examples in natural language.
\newblock In \emph{Proceedings of the 2020 Conference on Empirical Methods in
  Natural Language Processing: Findings}, pages 3829--3839.

\bibitem[{Morris et~al.(2020{\natexlab{b}})Morris, Lifland, Yoo, Grigsby, Jin,
  and Qi}]{morris2020textattack}
John Morris, Eli Lifland, Jin~Yong Yoo, Jake Grigsby, Di~Jin, and Yanjun Qi.
  2020{\natexlab{b}}.
\newblock Textattack: A framework for adversarial attacks, data augmentation,
  and adversarial training in nlp.
\newblock In \emph{Proceedings of the 2020 Conference on Empirical Methods in
  Natural Language Processing: System Demonstrations}, pages 119--126.

\bibitem[{Mozes et~al.(2021)Mozes, Stenetorp, Kleinberg, and
  Griffin}]{mozes2021frequency}
Maximilian Mozes, Pontus Stenetorp, Bennett Kleinberg, and Lewis Griffin. 2021.
\newblock Frequency-guided word substitutions for detecting textual adversarial
  examples.
\newblock In \emph{Proceedings of the 16th Conference of the European Chapter
  of the Association for Computational Linguistics: Main Volume}, pages
  171--186.

\bibitem[{Mrk{\v{s}}i{\'c} et~al.(2016)Mrk{\v{s}}i{\'c}, S{\'e}aghdha, Thomson,
  Ga{\v{s}}i{\'c}, Rojas-Barahona, Su, Vandyke, Wen, and
  Young}]{mrkvsic2016counter}
Nikola Mrk{\v{s}}i{\'c}, Diarmuid~O S{\'e}aghdha, Blaise Thomson, Milica
  Ga{\v{s}}i{\'c}, Lina Rojas-Barahona, Pei-Hao Su, David Vandyke, Tsung-Hsien
  Wen, and Steve Young. 2016.
\newblock Counter-fitting word vectors to linguistic constraints.
\newblock \emph{arXiv preprint arXiv:1603.00892}.

\bibitem[{Pedregosa et~al.(2011)Pedregosa, Varoquaux, Gramfort, Michel,
  Thirion, Grisel, Blondel, Prettenhofer, Weiss, Dubourg, Vanderplas, Passos,
  Cournapeau, Brucher, Perrot, and Duchesnay}]{scikit-learn}
F.~Pedregosa, G.~Varoquaux, A.~Gramfort, V.~Michel, B.~Thirion, O.~Grisel,
  M.~Blondel, P.~Prettenhofer, R.~Weiss, V.~Dubourg, J.~Vanderplas, A.~Passos,
  D.~Cournapeau, M.~Brucher, M.~Perrot, and E.~Duchesnay. 2011.
\newblock Scikit-learn: Machine learning in {P}ython.
\newblock \emph{Journal of Machine Learning Research}, 12:2825--2830.

\bibitem[{Princeton(2010)}]{wordnet}
Princeton. 2010.
\newblock Princeton university "about wordnet." wordnet.

\bibitem[{Pruthi et~al.(2019)Pruthi, Dhingra, and Lipton}]{pruthi2019combating}
Danish Pruthi, Bhuwan Dhingra, and Zachary~C Lipton. 2019.
\newblock Combating adversarial misspellings with robust word recognition.
\newblock In \emph{Proceedings of the 57th Annual Meeting of the Association
  for Computational Linguistics}, pages 5582--5591.

\bibitem[{Radford et~al.(2019)Radford, Wu, Child, Luan, Amodei, Sutskever
  et~al.}]{radford2019language}
Alec Radford, Jeffrey Wu, Rewon Child, David Luan, Dario Amodei, Ilya
  Sutskever, et~al. 2019.
\newblock Language models are unsupervised multitask learners.
\newblock \emph{OpenAI blog}, 1(8):9.

\bibitem[{Ren et~al.(2019)Ren, Deng, He, and Che}]{ren2019generating}
Shuhuai Ren, Yihe Deng, Kun He, and Wanxiang Che. 2019.
\newblock Generating natural language adversarial examples through probability
  weighted word saliency.
\newblock In \emph{Proceedings of the 57th annual meeting of the association
  for computational linguistics}, pages 1085--1097.

\bibitem[{Rousseeuw(1984)}]{rousseeuw1984least}
Peter~J Rousseeuw. 1984.
\newblock Least median of squares regression.
\newblock \emph{Journal of the American statistical association},
  79(388):871--880.

\bibitem[{Rousseeuw and Driessen(1999)}]{rousseeuw1999fast}
Peter~J Rousseeuw and Katrien~Van Driessen. 1999.
\newblock A fast algorithm for the minimum covariance determinant estimator.
\newblock \emph{Technometrics}, 41(3):212--223.

\bibitem[{Sch{\"o}lkopf et~al.(1997)Sch{\"o}lkopf, Smola, and
  M{\"u}ller}]{scholkopf1997kernel}
Bernhard Sch{\"o}lkopf, Alexander Smola, and Klaus-Robert M{\"u}ller. 1997.
\newblock Kernel principal component analysis.
\newblock In \emph{International conference on artificial neural networks},
  pages 583--588. Springer.

\bibitem[{Sch{\"o}lkopf et~al.(1998)Sch{\"o}lkopf, Smola, and
  M{\"u}ller}]{scholkopf1998nonlinear}
Bernhard Sch{\"o}lkopf, Alexander Smola, and Klaus-Robert M{\"u}ller. 1998.
\newblock Nonlinear component analysis as a kernel eigenvalue problem.
\newblock \emph{Neural computation}, 10(5):1299--1319.

\bibitem[{Socher et~al.(2013)Socher, Perelygin, Wu, Chuang, Manning, Ng, and
  Potts}]{socher-etal-2013-recursive}
Richard Socher, Alex Perelygin, Jean Wu, Jason Chuang, Christopher~D. Manning,
  Andrew Ng, and Christopher Potts. 2013.
\newblock \href {https://www.aclweb.org/anthology/D13-1170} {Recursive deep
  models for semantic compositionality over a sentiment treebank}.
\newblock In \emph{Proceedings of the 2013 Conference on Empirical Methods in
  Natural Language Processing}, pages 1631--1642, Seattle, Washington, USA.
  Association for Computational Linguistics.

\bibitem[{Wallace et~al.(2019)Wallace, Feng, Kandpal, Gardner, and
  Singh}]{wallace2019universal}
Eric Wallace, Shi Feng, Nikhil Kandpal, Matt Gardner, and Sameer Singh. 2019.
\newblock Universal adversarial triggers for attacking and analyzing nlp.
\newblock In \emph{Proceedings of the 2019 Conference on Empirical Methods in
  Natural Language Processing and the 9th International Joint Conference on
  Natural Language Processing (EMNLP-IJCNLP)}, pages 2153--2162.

\bibitem[{Wolf et~al.(2020)Wolf, Debut, Sanh, Chaumond, Delangue, Moi, Cistac,
  Rault, Louf, Funtowicz, Davison, Shleifer, von Platen, Ma, Jernite, Plu, Xu,
  Scao, Gugger, Drame, Lhoest, and Rush}]{wolf-etal-2020-transformers}
Thomas Wolf, Lysandre Debut, Victor Sanh, Julien Chaumond, Clement Delangue,
  Anthony Moi, Pierric Cistac, Tim Rault, Rémi Louf, Morgan Funtowicz, Joe
  Davison, Sam Shleifer, Patrick von Platen, Clara Ma, Yacine Jernite, Julien
  Plu, Canwen Xu, Teven~Le Scao, Sylvain Gugger, Mariama Drame, Quentin Lhoest,
  and Alexander~M. Rush. 2020.
\newblock \href {https://www.aclweb.org/anthology/2020.emnlp-demos.6}
  {Transformers: State-of-the-art natural language processing}.
\newblock In \emph{Proceedings of the 2020 Conference on Empirical Methods in
  Natural Language Processing: System Demonstrations}, pages 38--45, Online.
  Association for Computational Linguistics.

\bibitem[{Xie et~al.(2019)Xie, Wu, Maaten, Yuille, and He}]{xie2019feature}
Cihang Xie, Yuxin Wu, Laurens van~der Maaten, Alan~L Yuille, and Kaiming He.
  2019.
\newblock Feature denoising for improving adversarial robustness.
\newblock In \emph{Proceedings of the IEEE/CVF Conference on Computer Vision
  and Pattern Recognition}, pages 501--509.

\bibitem[{Xu et~al.(2018)Xu, Evans, and Qi}]{xu2017feature}
Weilin Xu, David Evans, and Yanjun Qi. 2018.
\newblock Feature squeezing: Detecting adversarial examples in deep neural
  networks.
\newblock \emph{Network and Distributed Systems Security Symposium}.

\bibitem[{Zhang et~al.(2020)Zhang, Sheng, Alhazmi, and
  Li}]{zhang2020adversarial}
Wei~Emma Zhang, Quan~Z Sheng, Ahoud Alhazmi, and Chenliang Li. 2020.
\newblock Adversarial attacks on deep-learning models in natural language
  processing: A survey.
\newblock \emph{ACM Transactions on Intelligent Systems and Technology (TIST)},
  11(3):1--41.

\bibitem[{Zhang et~al.(2015)Zhang, Zhao, and LeCun}]{zhang2015character}
Xiang Zhang, Junbo Zhao, and Yann LeCun. 2015.
\newblock Character-level convolutional networks for text classification.
\newblock \emph{Advances in neural information processing systems},
  28:649--657.

\bibitem[{Zhou et~al.(2021)Zhou, Zheng, Hsieh, Chang, and
  Huang}]{zhou2021defense}
Yi~Zhou, Xiaoqing Zheng, Cho-Jui Hsieh, Kai-Wei Chang, and Xuan-Jing Huang.
  2021.
\newblock Defense against synonym substitution-based adversarial attacks via
  dirichlet neighborhood ensemble.
\newblock In \emph{Proceedings of the 59th Annual Meeting of the Association
  for Computational Linguistics and the 11th International Joint Conference on
  Natural Language Processing (Volume 1: Long Papers)}, pages 5482--5492.

\bibitem[{Zhou et~al.(2019)Zhou, Jiang, Chang, and Wang}]{zhou2019learning}
Yichao Zhou, Jyun-Yu Jiang, Kai-Wei Chang, and Wei Wang. 2019.
\newblock Learning to discriminate perturbations for blocking adversarial
  attacks in text classification.
\newblock In \emph{Proceedings of the 2019 Conference on Empirical Methods in
  Natural Language Processing and the 9th International Joint Conference on
  Natural Language Processing (EMNLP-IJCNLP)}, pages 4904--4913.

\end{thebibliography}
\bibliographystyle{acl_natbib}

\clearpage
\appendix
\counterwithin{figure}{section}
\counterwithin{table}{section}
\section{Appendix}
\label{sec:appendix}
\subsection{Success Rate of Attack Methods and Other Statistics}
\label{appendix:attack description}
Here we briefly describe each attack method and provide some statistics about the attack results. 
For Table \ref{table:attack}, word transformation method indicates how candidate replacement words are created. Word importance ranking denotes how the ordering of which word to attack is chosen. For constraints, only those related to embedding was listed and the numbers in parenthesis denote the threshold. Higher threshold signifies stronger constraint. For more details, we refer the readers to \citet{morris2020textattack}. Table \ref{table:attack_results} summarizes the attack results on three dataset for BERT. Results for other models can be found in the released dataset.

\begin{table*}
\centering
{\small %
\begin{tabular}{llccc}
\hline
\textbf{Attacks} & \textbf{Citations} & \textbf{\makecell{Word Transformation\\ Method}} & \textbf{\makecell{Word Importance\\Ranking Method}} & \textbf{Constraints}\\
\hline
TF & 204 & \makecell{Counter-fitted GLOVE \\ \citep{mrkvsic2016counter}} & Delete Word  & \makecell{USE(0.84) \\ WordEmbedding Distance(0.5)}\\
\hline
PWWS & 187 & \makecell{WordNet\\ \citep{wordnet}} & Weighted Saliency & -\\
\hline 
BAE & 71 & Bert Masked LM & Delete Word & USE(0.94) \\
\hline
TF-adj & 18 & Counter-fitted GLOVE & Delete Word & \makecell{USE(0.98) \\ WordEmbedding Distance(0.9)}  \\
\hline
\end{tabular}}
\caption{Summary of attack methods and their defining characteristics.}
\label{table:attack}
\end{table*}

\begin{table}
{\small %
\begin{tabular}{lccc}
\hline
{\footnotesize \textbf{Attacks}} & {\scriptsize \textbf{\makecell{Post-Attack\\ Accuracy}}} & {\scriptsize \textbf{\makecell{Attack\\Success Rate}}} & {\scriptsize \textbf{\makecell{Average\\Num. Queries}}} \\
\hline
\multicolumn{4}{c}{IMDB (91.9\%)}\\
\hline
TF & 0.6\% & 99\% & 558 \\
PWWS & 3\% & 97\%  & 1681 \\
BAE & 34\% & 64\% & 455 \\
TF-adj & 84.2\% & 11\% & 298 \\
\hline
\multicolumn{4}{c}{AG-News (94.2\%)}\\
\hline
TF & 18\% & 81\% & 334 \\
PWWS & 41\% & 57\%  & 362 \\
BAE & 82\% & 14\% & 122 \\
TF-adj & 91\%& 5\% & 56 \\
\hline
\multicolumn{4}{c}{SST-2 (92.43\%)}\\
\hline
TF & 4\% & 96\% & 91 \\
PWWS & 12\% & 87\%  & 143 \\
BAE & 37\% & 61\% & 60 \\
TF-adj & 89\% & 5\% & 25 \\
\hline
\end{tabular}}
\caption{Summary of attack results for BERT on three datasets. Original accuracy of each dataset is written in parenthesis next to the dataset.}
\label{table:attack_results}
\end{table}

\begin{table*}[t]
\caption{Successful and Failed Adv. Examples in SST2 Dataset (\advsub{original}{replaced})}
\label{tab:example}
\small
\begin{tabularx}{1.0\textwidth}{lX|c}
\toprule
\textsc{Successful} & director brian levant, who never strays far from his sitcom roots, skates blithely from one \advsub{implausible}{improbable} situation to another [..] &  \includegraphics[height=0.4cm]{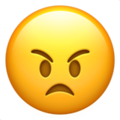} $\rightarrow$ \includegraphics[height=0.4cm]{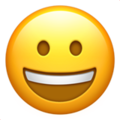}    \\ \midrule
\textsc{Failed} &  arnold 's \advsub{jump}{leap} from little screen to big will leave frowns on more than a few faces & \includegraphics[height=0.4cm]{figures/emoji/angry-face-emoji.png} $\rightarrow$ 
\includegraphics[height=0.4cm]{figures/emoji/angry-face-emoji.png} \\ \bottomrule
\end{tabularx}
\end{table*}

\subsection{Potential Errors of Parameter Estimation}
\label{appendix:parameter estimation}

\begin{figure}
\centering
\includegraphics[width=\columnwidth]{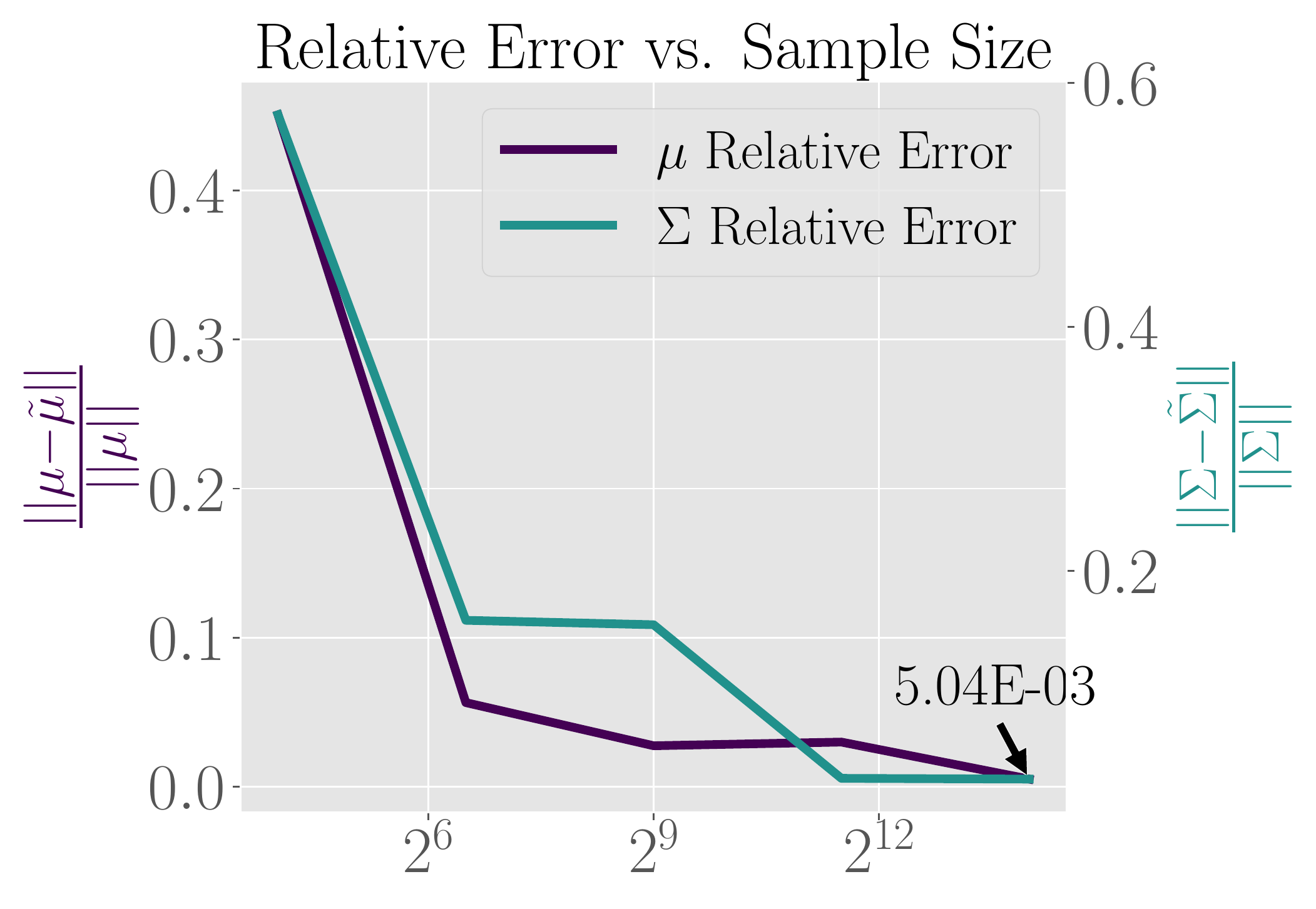}
\caption{Left figure shows the relative error of estimating the parameters of 768-dimensional multivariate Gaussian on a toy example. Even with $2^{14}$ samples, the relative error of $\mu$ is on the scale of $10^{-3}$}
\label{fig:relative_error}
\end{figure}
Accurate estimation of the parameters is difficult with finite amount of samples especially in high dimensions. Here we demonstrate this through a toy example and derive its relationship with the Mahalanobis distance function, which is proportional to the likelihood. Figure \ref{fig:relative_error} shows that the MLE error remains high even when $2^{14}$ samples are used to find the parameters of a noise-free normal distribution for both $\mu$ and $\Sigma$. This, in turn, leads to an inevitably error-prone $\tilde\mu=\mu-\epsilon_{\mu}$ and $\tilde\Delta = z-\tilde\mu$. Moreover, the error is amplified when computing the Mahalanobis distance due to the ill-conditioned $\tilde\Sigma$ with very small eigenvalues, which is observed empirically in all datasets and models (Figure \ref{fig:curse1}) possibly due to the redundant features. The (relative) condition number of the Mahalanobis distance function $g(\Delta)$ - relative change in the output given a relative change in the inputs - is bounded by the inverse of the smallest eigenvalue of $\tilde\Sigma^{-1}$. 
\begin{equation}
    \begin{aligned}
    \kappa_{g}(\Delta) &= \frac{|| \frac{\partial g}{\partial \Delta}||}{||g(\Delta)|| / ||\Delta||}\\ 
    &= \frac{||\Delta||}{||g(\Delta)||}||2 \Sigma^{-1}\Delta||\\
    &\leq \frac{||\Delta||}{||g(\Delta)||}2||\Sigma^{-1}|| ||\Delta||
    \end{aligned}
\end{equation}
where the first equality follows from the definition of condition number and differentiability of $g$ and $C_{\Delta}$. The last equality follows from the Caucy-Schwarz Inequality.
The matrix norm induced by the L2 norm is given by the largest singular value (largest eigenvalue for a positive definite square matrix). Given the eigenspectrum of $\Sigma$ as $\lambda_{\text{max}} \geq \dots \geq \lambda_{\text{min}}$, the eigenspectrum of $\Sigma^{-1}$ is given by the reciprocal of that of $\Sigma.$
Thus, $||\Sigma^{-1}||$ is equal to inverse of the minimum eigenvalue of $\Sigma$ and the last equality can be further decomposed into 
\begin{equation}
    \begin{aligned}
    \kappa_{g}(\Delta) &\leq \frac{||\Delta||}{||g(\Delta)||}2||\Sigma^{-1}|| ||\Delta||\\
    &\leq C_{\Delta} \frac{1}{\lambda_{\text{min}}}
    \end{aligned}
\end{equation}

where $C_{\Delta}$ is a constant for a given $\Delta$. This means that when the smallest eigenvalue is in the scale of $10^{-12}$, even a estimation error of scale $10^{-3}$ on $\mu$ may be amplified by at most by a scale of $10^9$. This leads to a serious problem in density estimation of $z$.

\subsection{More details on MCD}
\label{appendix:MCD}
We explain some of the properties of the determinant of the covariance matrix. First, the determinant is directly related to the differential entropy of the Gaussian distribution. For a $D$-dimensional multivariate Gaussian variable $X$ and its probability density function $f$, the differential entropy is given by  

\begin{equation}
    \begin{aligned}
    H(X) & = - \int_{\mathcal{X}} f(x) \log f(x) dx\\ 
         & = \frac{1}{2}\log ((2\pi{\rm e})^{D} {\rm det}(\Sigma)\\
         & \propto {\rm det}(\Sigma)
    \end{aligned}
\end{equation}

In addition, the determinant is also proportional to the volume of the ellipsoid for some $k$, $\{z \in \mathbb{R}^{D}: (z-\mu)^T \Sigma^{-1} (z-\mu)=k^{2}\}$. We refer the readers to Section 7.5 of \citet{anderson1962introduction} for the proof.  
This explain why the MCD estimate forms a much narrower probability contour than MLE as shown in Fig. \ref{fig:curse2}.

\SetKwInput{KwInput}{Input}              
\SetKwInput{KwOutput}{Output}  
\maketitle
\begin{algorithm}[t]
\DontPrintSemicolon
    \KwInput{$\mathcal{X}_{\text{train}}, \mathcal{Y}_{\text{train}}$, $\mathcal{D}=\{\mathcal{D}_{\text{adv}},\mathcal{D}_{\text{clean}}\}$}
    \KwInput{Feature Extractor $\mathcal{H}$}
    \KwOutput{Likelihood L}
    $\mathcal{Z}_{\text{train}}$ = $\mathcal{H}(\mathcal{X}_{\text{train}})$\;

    \If{MLE}
        {\For{\texttt{c in Class}}
        {
        $\tilde \mu_c= \frac{1}{N_c}\sum_{i\in Y_c}^{N_c} z_i$\;
        $\tilde \Sigma_c= \frac{1}{N_c} \sum_{i\in Y_c}^{N_c} (z_i-\tilde\mu) (z_i-\tilde\mu)^T$\;
        }
        }
    \ElseIf{RDE}
        {
        $\overline{Z}$ = kPCA($\mathcal{Z}$)\;
        \For{\text{c in Class}}
        {$\tilde\mu_c, \tilde\Sigma_c=$ MCD($\overline{Z_c}$)\;}
        }
    L = []\;
    \For{\texttt{$x$ in $\mathcal{D}$}}
        {
        $z= \mathcal{H}(x)$\;
        $\hat y= argmax_{k}$ $\mathcal{F}(x)_k$\;
        \If{RDE}
            {
            $z$ = kPCA(z)\;
            }
        L.\texttt{append}($\mathcal{N}(z|\tilde\mu_{\hat y},\tilde\Sigma_{\hat y})$)
        }
\caption{RDE and MLE}
\label{alg}
\end{algorithm}

\subsection{Implementation Details}
\label{appendix:implementation detail}
To meet the memory constraint of computing the kernel matrix, we sample a subset of $\mathcal{X}_{\text{train}}$ (8,000 samples) for all experiments. All models are pre-trained models provided by TextAttack and both kPCA and MCD are implemented using scikit-learn \citep{scikit-learn}.  We use the radial basis function as our kernel. 

For subsampling generated attacks described in Section \ref{subsec2.2:detecting}, we set the maximum number of adversarial samples for each dataset. For IMDB and AG-News, the maximum is set to 2000 and for SST-2 this is set to 1000. Then, the number of target samples (i.e. $||\mathcal{S}||\text{ or }||\mathcal{S}_1||$) is initialized to the maximum number divided by adversarial success ratio and task accuracy. Target sample is decremented until ratio between clean and adversarial samples can roughly be 5:5. Algorithm of RDE and MLE is provided in Algorithm \ref{alg}. 

\vfill
\subsection{Benchmark Usage Example}
\label{appendix:code}
\begin{lstlisting}
from AttackLoader import Attackloader

#Set seed, scenario, model type, 
#attack type, dataset, etc.
loader = AttackLoader(...)

#Split test and validation set 
loader.split_csv_to_testval()

# Subsample from testset according to chosen scenario 
sampled, _ = loader.get_attack_from_csv(..)

"""
Apply detection method 
"""
\end{lstlisting}

\begin{figure*}[!t]
\vspace{-3mm}
\centering
\includegraphics[width=1.0\textwidth]{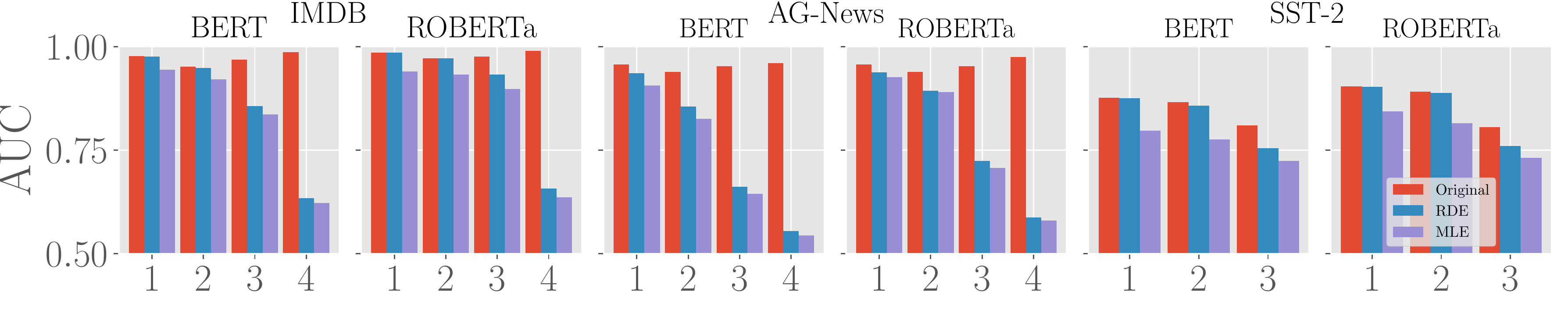}
\caption{Detection performance (\textsc{auc}) for scenario including failed adversarial samples. Horizontal axis dentoes the type of attacks methods (TF, PWWS, BAE, TF-adj). The original performance of RDE from Table \ref{tab:scenario1} as an upper bound is provided (red). Blue and purple denotes RDE and MLE including failed adversarial samples, respectively.}
\label{fig:fae}
\vspace{-4mm}
\end{figure*}

\subsection{Comparison with PCA}
\label{appendix:PCA}
In all our experiments, we used a radial basis function for kPCA. This allows finding non-linear patterns in the feature space. When the linear kernel is used, kPCA is equivalent to ordinary PCA. We demonstrate that exploiting non-linearity preserve much more meaningful information by comparing the detection performance in the IMDB dataset (Table \ref{tab:PCA}).

\subsection{ROC Curve Examples}
\label{appendix:ROC Curve}
\begin{figure*}
\includegraphics[width=\textwidth]{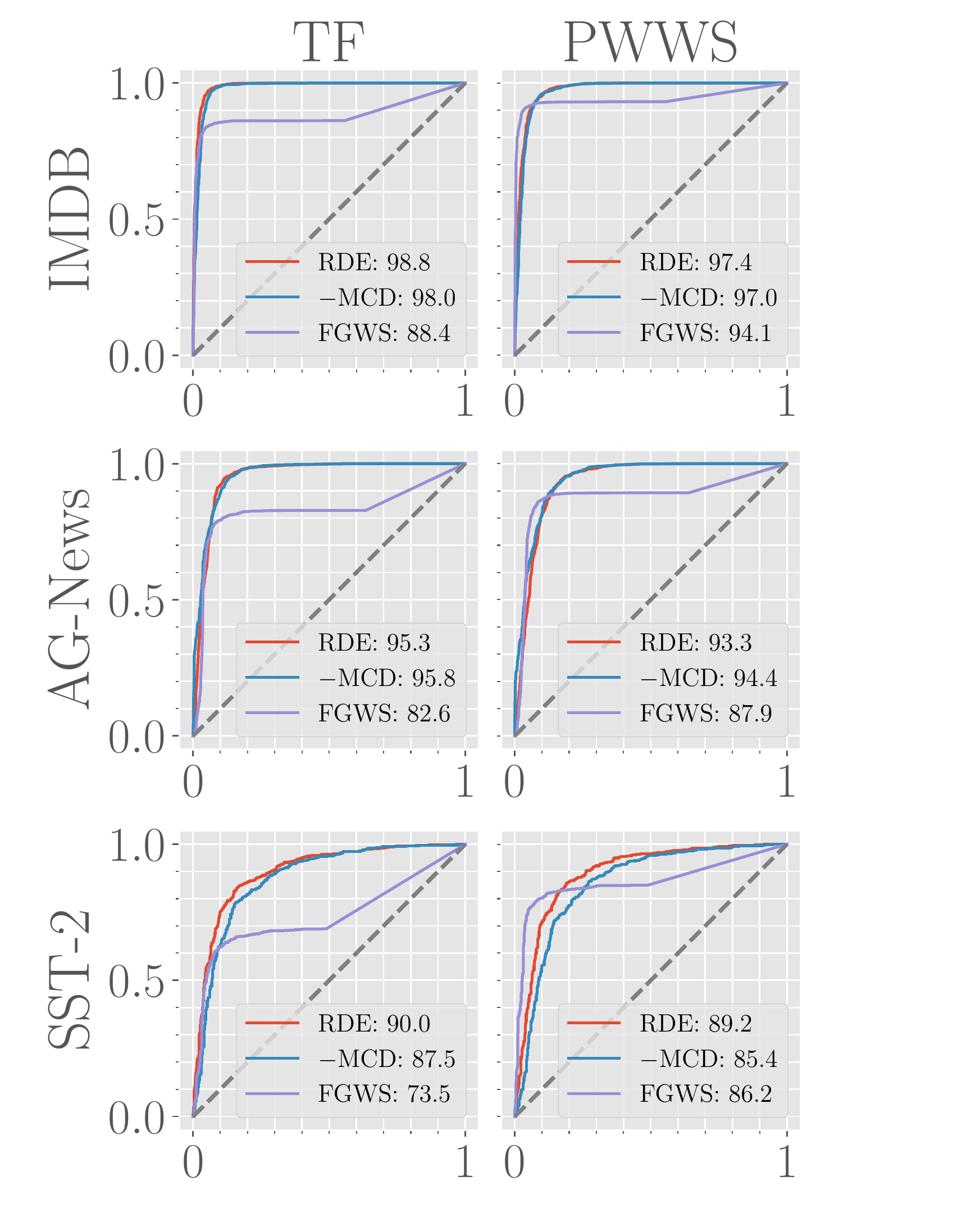}
\caption{ROC curves for RoBERTa on two attacks TF and PWWS. Row indicates the dataset, while the column indicates the attack methods. For all plots, the x-axis and y-axis represents \textsc{fpr} and \textsc{tpr}, respectively. The legend indicates the \textsc{auc} of each methods. RDE(-MCD) is written as -MCD due to space constraint.}
\label{fig:ROC}
\end{figure*}

Below (Fig. \ref{fig:ROC}) we provide Receiver Operating Characteristic (ROC) curves of RoBERTa on two attacks. For all plots, samples from the first seed are used.

\subsection{Experiment Details on Adaptive Adversary}
\label{appendix:adaptive}
In this section, we provide detail on the experimental settings. Instead of the terminating condition used by all attack methods
\begin{equation}
\begin{gathered}
    \label{Eq1: adversarial example}
    \mathcal{F^{*}}(x) = \mathcal{F}(x) \neq \mathcal{F}(x_\text{adv})
\end{gathered}
\end{equation}
the stronger attack goes on to send the sample closer to the features of the incorrect target. With fixed $\epsilon$, the attack terminates once
\begin{equation}
\begin{gathered}
    \label{Eq1: adversarial example}
    \mathcal{F^{*}}(x) = \mathcal{F}(x) \neq \mathcal{F}(x_\text{adv}),\\
    \hat{p} > 1-\epsilon
\end{gathered}
\end{equation}
where $\hat{p}$ is the largest predicted softmax probability. 
This allows the generated sample to fool the density-based methods that rely on the discriminative feature space. For attacks denoted 'strong' $\epsilon=0.1$; for 'stronger', $\epsilon=0.01$. For RDE+, we use the adversarial samples of the held-out validation set using the static attack method.

\subsection{Detecting Character-level Attacks}
\label{appendix:char-level}
Although character level attacks are perceptible to spell checkers or more sophisticated techniques \citep{pruthi2019combating}, it still poses threat to deep neural networks \citep{zhang2020adversarial}. We demonstrate the general applicability of RDE on charcter-level attack on 3,000 samples attacked by the character-level attack method proposed in \citet{pruthi2019combating}. As shown in Table \ref{table:char-level}, RDE surpasses all the density-based methods.

\begin{table}[h!]
\begin{adjustbox}{width=\columnwidth,center}
\begin{tabular}{c||cc|cc}
\hline
        & \multicolumn{2}{c}{IMDB}         & \multicolumn{2}{c}{AG-News}  \\
        & \textbf{BERT} & \textbf{RoBERTa} & \textbf{BERT} & \textbf{RoBERTa} \\ 
    PPL & 55.0$\pm$0.6  &  64.0$\pm$1.4    &  71.3$\pm$0.7 & 71.3$\pm$0.7 \\
    MLE & 90.2$\pm$0.4  &  89.6$\pm$0.3    &  89.9$\pm$0.4 & 77.5$\pm$0.6 \\
    RDE & 91.0$\pm$0.1  &  92.9$\pm$0.2    &  90.9$\pm$0.5 & 91.0$\pm$0.4 \\
    
\hline
\end{tabular}
\end{adjustbox}
\caption{\textsc{AUC} for character level attack on AG-News averaged over five trials.}
\label{table:char-level}
\end{table}

\subsection{More Realistic Scenarios}
\label{appendix:scenario2}
\noindent In this section, we discuss more realistic scenarios and provide preliminary results in Appendix \ref{appendix:scenario2}: (i) Including failed attacks (ii) Imbalanced ratio of clean and adversarial samples. In previous experiments, all failed adversarial attacks were discarded. However, in reality an adversary will probably have no access to the victim model so some attacks will indeed fail to fool the model. While failed adversarial attacks do not pose threat to the task accuracy of the model, it nevertheless may be harmful if the victim wishes to gain information about a certain population by aggregating data such as sentiment in consumer review about a movie.
In addition, as active research in attacks have been made in the past few years, more subtle attacks that are imperceptible to humans naturally have lower attack success ratio (e.g. BAE).  

Fig. \ref{fig:fae} demonstrates the detection results of RDE and MLE when distinguishing between normal samples and (failed and successful) adversarial attempts by comparing the \textsc{auc}'s. As an upper bound, we provide the performance of RDE on the \textit{original scenario} without failed adversarial examples in red. As the first two attacks (TF and PWWS) achieve nearly 100\% success rate, only few failed adversarial samples are added. Accordingly, the performances for the two attacks show little difference. However, in more subtle attacks (BAE and TF-adj) the performance drastically drops due to the increased failed adversarial samples, yet RDE outperforms MLE by a considerable margin in most cases. We end on this topic by noting that more comprehensive analysis is called for, because in some cases failed adversarial attempts are (nearly) identical to clean samples. So an attack method with low detection rate does not necessarily imply a crafty attack method in this scenario. 

In Appendix Table \ref{tab:scenario2}, we provide the results for Scenario 2 described in Section \ref{subsec2.2:detecting}. The general trend among detection methods and attack methods is similar to Table \ref{tab:scenario1}. As noted earlier, for Scenario 2 the ratio of adversarial to clean samples will be low if the attack success rate is low. For instance, in IMDB-(TF-adj)-BERT, the ratio of adversarial to clean samples is around 1:9. Whereas both \textsc{auc} and \textsc{tpr} are not strongly affected due to the characteristic of the metrics, F1 drastically drops. For instance, for IMDB-(TF-adj)-BERT, RDE achieves 73.7\% (as opposed to 95.0\% of Scenario 1). On the same set, FGWS achieves 60.1\% and MLE achieves 67.6\%. 

Here we provide experimental results on Scenario 2. Although pairs of samples are included in the dataset, the general trend is similar to that of Scenario 1. For attack methods with low success rate, the adversarial to clean sample ratio is low, which affects the F1-score.

\begin{table*}
\begin{adjustbox}{width=1.0\textwidth}
\begin{tabular}{c|c||c|c|c||c|c|c||c|c|c||c|c|c}
 \hline
 \multirow{3}{*}{Models} & \multirow{3}{*}{Methods} & \multicolumn{12}{c}{Attacks}\\
 \cline{3-14}
 && \multicolumn{3}{c||}{TF} & \multicolumn{3}{c||}{PWWS}& \multicolumn{3}{c||}{BAE} & \multicolumn{3}{c}{TF-adj}\\
 \cline{3-14}
 && \textsc{tpr} & F1 & \textsc{auc} & \textsc{tpr} & F1 & \textsc{auc} & \textsc{tpr} & F1 & \textsc{auc} & \textsc{tpr} & F1 & \textsc{auc} \\
 \hline
     \multirow{2}{*}{{BERT}} 
    & PCA &89.5$\pm$1.1 & 89.7$\pm$0.6& 95.2$\pm$0.1&77.9$\pm$1.3 & 82.9$\pm$0.8& 93.1$\pm$0.2&83.5$\pm$1.3 & 86.3$\pm$0.7& 94.2$\pm$0.2&92.8$\pm$1.2 & 91.7$\pm$0.7& 96.2$\pm$0.2 \\ 
    & kPCA & 96.3$\pm$0.3 & 93.4$\pm$0.2& 96.8$\pm$0.1&86.9$\pm$0.9 & 88.3$\pm$0.5& 94.6$\pm$0.2&92.5$\pm$0.5 & 91.4$\pm$0.3& 95.8$\pm$0.2&98.2$\pm$0.2 & 94.6$\pm$0.2& 97.6$\pm$0.2 \\ 
    \cmidrule{1-14}\morecmidrules\cmidrule{1-14}
    \multirow{2}{*}{{\small RoBERTa}} 
    & PCA &94.7$\pm$0.6 & 92.5$\pm$0.3& 96.3$\pm$0.1&89.7$\pm$0.9 & 89.9$\pm$0.5& 95.4$\pm$0.1&88.2$\pm$1.3 & 89.0$\pm$0.7& 95.0$\pm$0.3&92.6$\pm$1.8 & 91.6$\pm$0.9& 96.7$\pm$0.5 \\ 
    & kPCA &98.5$\pm$0.1 & 94.5$\pm$0.1& 97.9$\pm$0.1&95.0$\pm$0.3 & 92.7$\pm$0.2& 96.7$\pm$0.1& 95.4$\pm$0.4 &93.0$\pm$0.2& 97.0$\pm$0.2&98.6$\pm$0.4 & 94.8$\pm$0.2& 98.1$\pm$0.4 \\ 
\end{tabular}
\end{adjustbox}
\caption{Results of using linear kernel for KPCA, which is equivalent to the ordinary PCA on IMDB. For fair comparison, we compare with RDE(-MCD) where MCD estimation is not used. All results lead to higher performance when kPCA is used.}
\label{tab:PCA}
\end{table*}

\begin{table*}
\begin{adjustbox}{width=1.0\textwidth}
\begin{tabular}{c|c||c|c|c||c|c|c||c|c|c||c|c|c}
 \hline
 \multirow{3}{*}{Models} & \multirow{3}{*}{Methods} & \multicolumn{12}{c}{Attacks}\\
 \cline{3-14}
 && \multicolumn{3}{c||}{TF} & \multicolumn{3}{c||}{PWWS}& \multicolumn{3}{c||}{BAE} & \multicolumn{3}{c}{TF-adj}\\
 \cline{3-14}
 && \textsc{tpr} & F1 & \textsc{auc} & \textsc{tpr} & F1 & \textsc{auc} & \textsc{tpr} & F1 & \textsc{auc} & \textsc{tpr} & F1 & \textsc{auc} \\
 \hline 
 \cline{3-14}
 \multirow{5}{*}{{BERT}}  
    & PPL &40.2$\pm$0.2 & 53.5$\pm$0.2& 76.6$\pm$0.0&36.2$\pm$0.7 & 49.6$\pm$0.8& 74.2$\pm$0.4&16.3$\pm$0.7 & 25.9$\pm$0.9& 65.2$\pm$0.6&16.3$\pm$1.5 & 25.7$\pm$2.0& 63.1$\pm$0.6 \\ 
    & FGWS &84.6$\pm$0.5 & 87.2$\pm$0.2& 87.0$\pm$0.5 &88.9$\pm$0.1 & 89.5$\pm$0.1& 91.0$\pm$0.0& 62.4$\pm$1.2 & 72.5$\pm$0.8& 70.3$\pm$0.8&79.7$\pm$2.7 & 85.5$\pm$1.7& 82.3$\pm$0.8 \\ 
    & MLE &41.3$\pm$1.0 & 54.6$\pm$1.0& 66.5$\pm$0.3&41.8$\pm$0.8 & 55.0$\pm$0.8& 66.9$\pm$0.2&42.3$\pm$1.9 & 55.5$\pm$1.8& 67.5$\pm$0.4&38.5$\pm$2.1 & 52.0$\pm$2.0& 66.7$\pm$0.8 \\ 
    & RDE &\textbf{97.9}$\pm$0.1 & \textbf{94.3}$\pm$0.1& \textbf{96.}4$\pm$0.0&\textbf{92.7}$\pm$0.5 & \textbf{91.5}$\pm$0.3& \textbf{94.9}$\pm$0.1&\textbf{97.1}$\pm$0.2 &\textbf{ 94.0}$\pm$0.1&\textbf{ 96.2}$\pm$0.1&\textbf{99.8}$\pm$0.1 & \textbf{95.7}$\pm$0.1& \textbf{96.4}$\pm$0.1 \\ 
    \cmidrule{1-14}\morecmidrules\cmidrule{1-14}
    \multirow{5}{*}{{\small RoBERTa}} 
    & PPL &38.5$\pm$0.7 & 51.8$\pm$0.7& 73.0$\pm$0.5&36.3$\pm$0.8 & 49.6$\pm$0.9& 71.7$\pm$0.6&17.0$\pm$0.6 & 26.7$\pm$0.7& 60.6$\pm$0.7&12.7$\pm$1.2 & 20.8$\pm$1.7& 58.6$\pm$0.5 \\ 
    & FGWS &83.5$\pm$0.1 & 86.5$\pm$0.1& 86.6$\pm$0.1&86.8$\pm$0.2 & 88.3$\pm$0.1& 89.7$\pm$0.1&61.9$\pm$0.6 & 72.6$\pm$0.4& 70.9$\pm$0.4&69.1$\pm$2.7 & 78.7$\pm$1.7& 75.6$\pm$2.3 \\ 
    & MLE &95.0$\pm$0.4 & 92.7$\pm$0.2& 96.3$\pm$0.1&89.2$\pm$0.4 & 89.6$\pm$0.2& 94.5$\pm$0.2&90.1$\pm$0.3 & 90.1$\pm$0.2& 94.9$\pm$0.2&85.7$\pm$3.5 & 87.6$\pm$2.0& 95.0$\pm$0.5 \\ 
    & RDE & \textbf{96.4}$\pm$0.1 & \textbf{93.5}$\pm$0.0& \textbf{97.0}$\pm$0.1&\textbf{90.7}$\pm$0.1 & \textbf{90.5}$\pm$0.1& \textbf{95.3}$\pm$0.0& \textbf{92.5}$\pm$0.2 & \textbf{91.5}$\pm$0.1& \textbf{95.7}$\pm$0.2& \textbf{99.5}$\pm$0.3 & \textbf{95.5}$\pm$0.0& \textbf{96.5}$\pm$0.5 \\ 
 \hline
\end{tabular}
\end{adjustbox}
\caption{Experiment results on the Yelp dataset}
\label{tab:yelp}
\end{table*}

\begin{table*}[hb]
\begin{adjustbox}{width=\textwidth}
\begin{tabular}{c|c||c|c|c||c|c|c||c|c|c||c|c|c}
 \hline
 \multirow{3}{*}{Models} & \multirow{3}{*}{Methods} & \multicolumn{12}{c}{Attacks}\\
 \cline{3-14}
 && \multicolumn{3}{c||}{TF} & \multicolumn{3}{c||}{PWWS}& \multicolumn{3}{c||}{BAE} & \multicolumn{3}{c}{TF-adj}\\
 \cline{3-14}
 && \textsc{tpr} & F1 & \textsc{auc} & \textsc{tpr} & F1 & \textsc{auc} & \textsc{tpr} & F1 & \textsc{auc} & \textsc{tpr} & F1 & \textsc{auc} \\
 \hline
 && \multicolumn{12}{c}{\Large IMDB} \\ 
 \hline
 \multirow{5}{*}{{BERT}} 
    & PPL &49.3$\pm$0.2 & 61.5$\pm$0.2& 77.4$\pm$0.2&38.9$\pm$0.3 & 51.9$\pm$0.3& 71.9$\pm$0.2&28.1$\pm$0.3 & 38.8$\pm$0.3& 67.6$\pm$0.1&24.3$\pm$0.1 & 25.0$\pm$0.2& 66.7$\pm$0.1 \\ 
    & FGWS &82.6$\pm$0.1 & 85.4$\pm$0.1& 85.1$\pm$0.1&86.6$\pm$0.1 & 87.7$\pm$0.1& 89.3$\pm$0.1&60.6$\pm$0.2 & 68.5$\pm$0.2& 69.3$\pm$0.2&71.3$\pm$0.2 & 60.1$\pm$0.2& 76.6$\pm$0.1 \\ 
    & MLE &86.4$\pm$1.2 & 87.6$\pm$0.7& 94.6$\pm$0.1&76.1$\pm$1.7 & 81.3$\pm$1.1& 92.6$\pm$0.2&82.0$\pm$0.6 & 82.6$\pm$0.3& 93.7$\pm$0.1&87.0$\pm$0.2 & 67.6$\pm$0.2& 95.0$\pm$0.0 \\
    \cline{2-14}
    & RDE(-MCD) &96.0$\pm$0.4 & 92.8$\pm$0.2& 96.6$\pm$0.1&86.2$\pm$0.8 & 87.5$\pm$0.5& 94.4$\pm$0.1&92.1$\pm$0.3 & 88.3$\pm$0.2& 95.6$\pm$0.1&98.5$\pm$0.0 & 73.4$\pm$0.0& 97.5$\pm$0.0 \\ 
    & RDE &96.8$\pm$0.2 & 93.2$\pm$0.1& 97.7$\pm$0.1&87.4$\pm$0.5 & 88.1$\pm$0.3& 95.1$\pm$0.2&93.3$\pm$0.3 & 89.0$\pm$0.1& 96.8$\pm$0.1&98.8$\pm$0.0 & 73.7$\pm$0.1& 98.6$\pm$0.0 \\ 

    \cmidrule{1-14}\morecmidrules\cmidrule{1-14}
    \multirow{5}{*}{{\small RoBERTa}} 
    & PPL &49.5$\pm$0.4 & 61.8$\pm$0.3& 78.9$\pm$0.3&45.1$\pm$0.2 & 57.9$\pm$0.2& 76.5$\pm$0.2&26.9$\pm$0.1 & 37.9$\pm$0.1& 67.6$\pm$0.1&26.6$\pm$0.4 & 21.3$\pm$0.3& 68.1$\pm$0.3 \\ 
    & FGWS &83.5$\pm$0.2 & 86.2$\pm$0.1& 86.6$\pm$0.2&91.6$\pm$0.1 & 90.7$\pm$0.0& 93.1$\pm$0.1&60.7$\pm$0.1 & 69.1$\pm$0.1& 69.3$\pm$0.2&66.0$\pm$0.4 & 46.9$\pm$0.2& 72.9$\pm$0.4 \\ 
    & MLE &80.7$\pm$0.8 & 84.4$\pm$0.5& 94.0$\pm$0.0&77.3$\pm$1.0 & 82.3$\pm$0.6& 93.2$\pm$0.1&75.9$\pm$1.0 & 79.5$\pm$0.6& 93.0$\pm$0.1&84.1$\pm$0.7 & 54.7$\pm$0.2& 94.3$\pm$0.0 \\
    \cline{2-14}
    & RDE(-MCD) &98.1$\pm$0.1 & 94.1$\pm$0.0& 97.9$\pm$0.0&94.7$\pm$0.3 & 92.3$\pm$0.2& 96.7$\pm$0.0&95.0$\pm$0.1 & 90.5$\pm$0.0& 96.8$\pm$0.0&98.9$\pm$0.1 & 61.7$\pm$0.1& 97.8$\pm$0.0 \\ 
    & RDE &98.5$\pm$0.1 & 94.3$\pm$0.0& 98.6$\pm$0.0&94.9$\pm$0.4 & 92.4$\pm$0.2& 97.2$\pm$0.0&94.8$\pm$0.1 & 90.4$\pm$0.0& 97.5$\pm$0.0&99.1$\pm$0.1 & 62.1$\pm$0.1& 98.9$\pm$0.0 \\ 

 \hline
 && \multicolumn{12}{c}{\Large AG-News} \\ 
 \hline
 \multirow{5}{*}{{BERT}}  
    & PPL &76.3$\pm$0.3 & 80.7$\pm$0.2& 91.3$\pm$0.1&72.4$\pm$0.4 & 75.8$\pm$0.3& 90.2$\pm$0.1&30.6$\pm$0.5 & 29.9$\pm$0.4& 72.8$\pm$0.2&35.0$\pm$0.5 & 19.8$\pm$0.2& 74.3$\pm$0.4 \\ 
    & FGWS &82.4$\pm$0.6 & 84.4$\pm$0.4& 84.2$\pm$0.7&90.9$\pm$0.2 & 86.8$\pm$0.1& 90.0$\pm$0.1&62.9$\pm$0.3 & 53.0$\pm$0.3& 70.5$\pm$0.1&66.0$\pm$0.6 & 36.3$\pm$0.6& 72.4$\pm$0.4 \\ 
    & MLE &78.2$\pm$0.4 & 81.8$\pm$0.3& 93.5$\pm$0.0&71.4$\pm$0.3 & 75.3$\pm$0.2& 92.3$\pm$0.1&69.9$\pm$0.3 & 57.4$\pm$0.1& 92.2$\pm$0.0&64.0$\pm$0.5 & 33.4$\pm$0.4& 91.1$\pm$0.1 \\
    \cline{2-14}
    & RDE(-MCD) &96.3$\pm$0.3 & 92.1$\pm$0.1& 97.2$\pm$0.0&90.5$\pm$0.3 & 86.6$\pm$0.1& 95.7$\pm$0.1&91.4$\pm$0.3 & 68.8$\pm$0.1& 95.5$\pm$0.0&92.2$\pm$0.5 & 44.4$\pm$0.1& 95.6$\pm$0.0 \\ 
    & RDE &96.0$\pm$0.3 & 91.9$\pm$0.1& 97.0$\pm$0.0&89.8$\pm$0.4 & 86.2$\pm$0.2& 95.4$\pm$0.0&94.0$\pm$0.2 & 69.9$\pm$0.1& 96.2$\pm$0.0&96.6$\pm$0.1 & 47.4$\pm$0.3& 96.6$\pm$0.0 \\ 

    \cmidrule{1-14}\morecmidrules\cmidrule{1-14}
    \multirow{5}{*}{{\small RoBERTa}} 
    & PPL &77.3$\pm$0.3 & 81.5$\pm$0.2& 91.8$\pm$0.1&73.0$\pm$0.3 & 77.2$\pm$0.2& 90.0$\pm$0.1&36.2$\pm$0.5 & 36.3$\pm$0.5& 74.9$\pm$0.3&36.2$\pm$0.3 & 21.5$\pm$0.2& 75.8$\pm$0.2 \\ 
    & FGWS &79.6$\pm$0.4 & 83.0$\pm$0.2& 82.6$\pm$0.3&86.6$\pm$0.3 & 85.5$\pm$0.2& 88.0$\pm$0.2&52.7$\pm$0.4 & 48.9$\pm$0.3& 64.6$\pm$0.5&60.7$\pm$0.7 & 34.4$\pm$0.3& 70.2$\pm$0.3 \\ 
    & MLE &81.4$\pm$0.2 & 84.1$\pm$0.1& 94.0$\pm$0.1&78.7$\pm$0.0 & 80.8$\pm$0.0& 93.1$\pm$0.0&68.4$\pm$0.2 & 59.1$\pm$0.2& 91.7$\pm$0.1&62.6$\pm$0.2 & 34.3$\pm$0.2& 90.0$\pm$0.0 \\
    \cline{2-14}
    & RDE(-MCD) &89.9$\pm$0.3 & 89.0$\pm$0.2& 96.1$\pm$0.0&85.8$\pm$0.5 & 85.0$\pm$0.3& 95.1$\pm$0.1&81.4$\pm$0.2 & 66.6$\pm$0.2& 94.4$\pm$0.1&80.4$\pm$0.1 & 42.0$\pm$0.2& 93.7$\pm$0.0 \\ 
    & RDE &92.7$\pm$0.2 & 90.5$\pm$0.1& 95.6$\pm$0.0&86.4$\pm$0.3 & 85.4$\pm$0.2& 94.2$\pm$0.1&92.6$\pm$0.2 & 72.3$\pm$0.2& 95.7$\pm$0.0&93.6$\pm$0.1 & 47.6$\pm$0.4& 95.7$\pm$0.0 \\ 

 \hline
 && \multicolumn{12}{c}{\Large SST-2} \\ 
 \hline
 \multirow{5}{*}{{BERT}}  
    & PPL &33.4$\pm$0.5 & 46.2$\pm$0.6& 73.1$\pm$0.1&30.4$\pm$0.5 & 42.6$\pm$0.5& 73.1$\pm$0.0&22.2$\pm$0.1 & 31.7$\pm$0.1& 65.7$\pm$0.0 \\ 
    & FGWS &61.4$\pm$0.6 & 71.2$\pm$0.5& 73.5$\pm$0.4&79.4$\pm$0.2 & 82.9$\pm$0.1& 86.2$\pm$0.1&33.0$\pm$0.3 & 43.8$\pm$0.3& 61.2$\pm$0.2 \\ 
    & MLE &33.3$\pm$0.2 & 46.1$\pm$0.2& 80.5$\pm$0.1&23.3$\pm$0.5 & 34.4$\pm$0.6& 78.4$\pm$0.1&34.1$\pm$0.0 & 45.0$\pm$0.0& 76.6$\pm$0.0 \\
    \cline{2-11}
    & RDE(-MCD) &60.5$\pm$0.6 & 70.6$\pm$0.4& 86.4$\pm$0.2&45.9$\pm$0.3 & 58.0$\pm$0.3& 83.8$\pm$0.1&44.1$\pm$0.0 & 54.5$\pm$0.0& 79.9$\pm$0.0 \\ 
    & RDE &66.3$\pm$0.3 & 74.8$\pm$0.2& 87.6$\pm$0.2&53.0$\pm$0.1 & 64.2$\pm$0.1& 85.8$\pm$0.1&47.6$\pm$0.1 & 57.7$\pm$0.1& 80.3$\pm$0.0 \\ 

    \cmidrule{1-11}\morecmidrules\cmidrule{1-11}
    \multirow{5}{*}{{\small RoBERTa}} 
    & PPL &35.1$\pm$0.2 & 48.1$\pm$0.2& 74.5$\pm$0.0&33.3$\pm$0.4 & 45.8$\pm$0.5& 74.0$\pm$0.1&21.2$\pm$0.1 & 30.6$\pm$0.2& 64.7$\pm$0.1 \\ 
    & FGWS &61.4$\pm$0.4 & 71.2$\pm$0.3& 73.7$\pm$0.2&80.1$\pm$0.3 & 83.4$\pm$0.2& 86.2$\pm$0.1&36.7$\pm$0.4 & 47.7$\pm$0.4& 60.5$\pm$0.1 \\ 
    & MLE &41.8$\pm$0.3 & 54.7$\pm$0.2& 84.2$\pm$0.1&31.6$\pm$0.4 & 44.0$\pm$0.4& 81.5$\pm$0.2&37.0$\pm$0.0 & 48.0$\pm$0.0& 77.7$\pm$0.0 \\
    \cline{2-11}
    & RDE(-MCD) &62.5$\pm$0.6 & 72.1$\pm$0.5& 87.5$\pm$0.3&51.9$\pm$0.6 & 63.3$\pm$0.5& 84.7$\pm$0.1&45.6$\pm$0.3 & 56.1$\pm$0.2& 79.2$\pm$0.1 \\ 
    & RDE &73.3$\pm$0.6 & 79.6$\pm$0.4& 90.4$\pm$0.2&65.7$\pm$0.3 & 73.9$\pm$0.2& 88.5$\pm$0.1&50.9$\pm$0.1 & 60.6$\pm$0.1& 80.2$\pm$0.1 \\ 

 \hline
\end{tabular}
\end{adjustbox}
\caption{Adversarial detection results for BERT and RoBERTa on Scenario 2 on three datasets (IMDB, AG-News, SST-2). For all three metrics, higher means better.}
\label{tab:scenario2}
\end{table*}

\end{document}